\documentclass[10pt,twocolumn,letterpaper]{article}
\usepackage{booktabs}
\usepackage{multirow}
\usepackage{mathtools, cases}
\usepackage{caption}
\usepackage{subcaption}
\usepackage{tabularx}
\usepackage{algpseudocode}
\usepackage{algorithm2e}

\usepackage[pagenumbers]{cvpr} 

\usepackage[dvipsnames]{xcolor}

\definecolor{cvprblue}{rgb}{0.21,0.49,0.74}
\usepackage[pagebackref,breaklinks,colorlinks,citecolor=cvprblue]{hyperref}

\def\paperID{10000} 
\def\confName{CVPR}
\def\confYear{2024}

\title{RoCoTex: A Robust Method for Consistent Texture Synthesis with Diffusion Models}

\author{
    Jangyeong Kim$^{1,2,\dagger}$ \and Donggoo Kang$^{1,3,\dagger}$ \and Junyoung Choi$^{1}$ \and Jeonga Wi$^{1}$ \and Junho Gwon$^{1}$ \and Jiun Bae$^{1}$ \and Dumim Yoon$^{1}$ \and Junghyun Han$^{2,*}$
    \and
    $^1$Graphics AI Lab, NCSOFT\\
    $^2$Department of Computer Science and Engineering, Korea University \\
    $^3$Department of Image, Chung-Ang University \\
    \tt\small \{jangyeongk, jychoi13, jaywi, jiunbae, yoondm\}@ncsoft.com, \\
    \tt\small tiruss@cau.ac.kr, dev@uos.ac.kr, jhan@korea.ac.kr
}

\begin{document}
\twocolumn[{
\maketitle
\begin{center}
    \centering
    \captionsetup{type=figure}
    \includegraphics[width=0.90\textwidth]{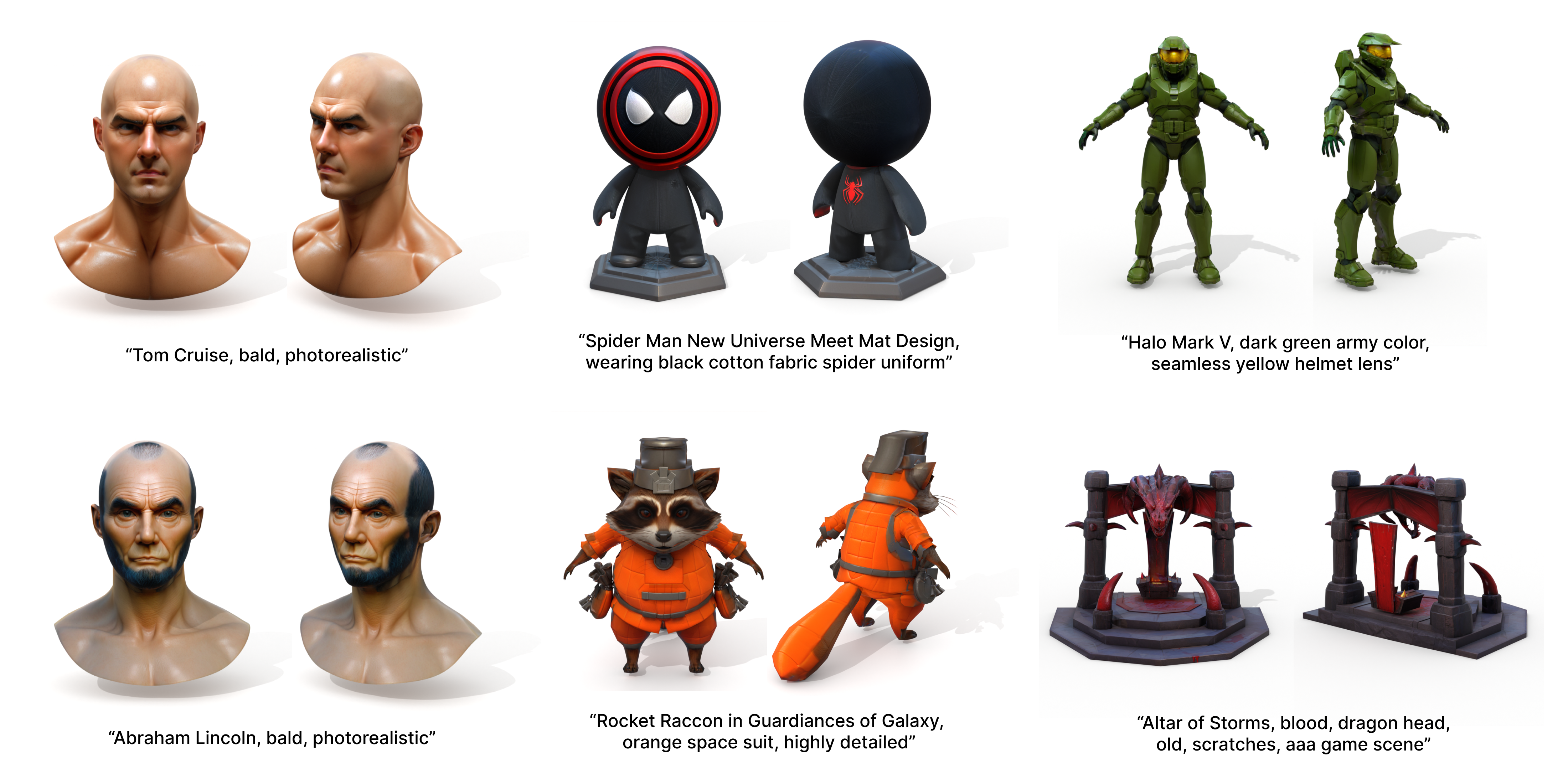}
    \captionof{figure}{This paper presents RoCoTex, a diffusion-based text-to-texture generation method that addresses the challenges of synthesizing view-consistent, well-aligned, seamless, and high-quality textures.}
\end{center}
}]

\begin{abstract}
Text-to-texture generation has recently attracted increasing attention, but existing methods often suffer from the problems of view inconsistencies, apparent seams, and misalignment between textures and the underlying mesh. In this paper, we propose a robust text-to-texture method for generating consistent and seamless textures that are well aligned with the mesh. Our method leverages state-of-the-art 2D diffusion models, including SDXL and multiple ControlNets, to capture structural features and intricate details in the generated textures.
The method also employs a symmetrical view synthesis strategy combined with regional prompts for enhancing view
consistency. Additionally, it introduces novel texture blending and soft-inpainting techniques, which significantly reduce the seam regions.
Extensive experiments demonstrate that our method outperforms existing state-of-the-art methods.
\end{abstract}


\section{Introduction}
\label{sec:intro}
\begin{figure*}[t]
    \centering
    \vspace{0.5cm}
    \begin{subfigure}{0.32\textwidth}
        \centering
        \includegraphics[width=1\textwidth]{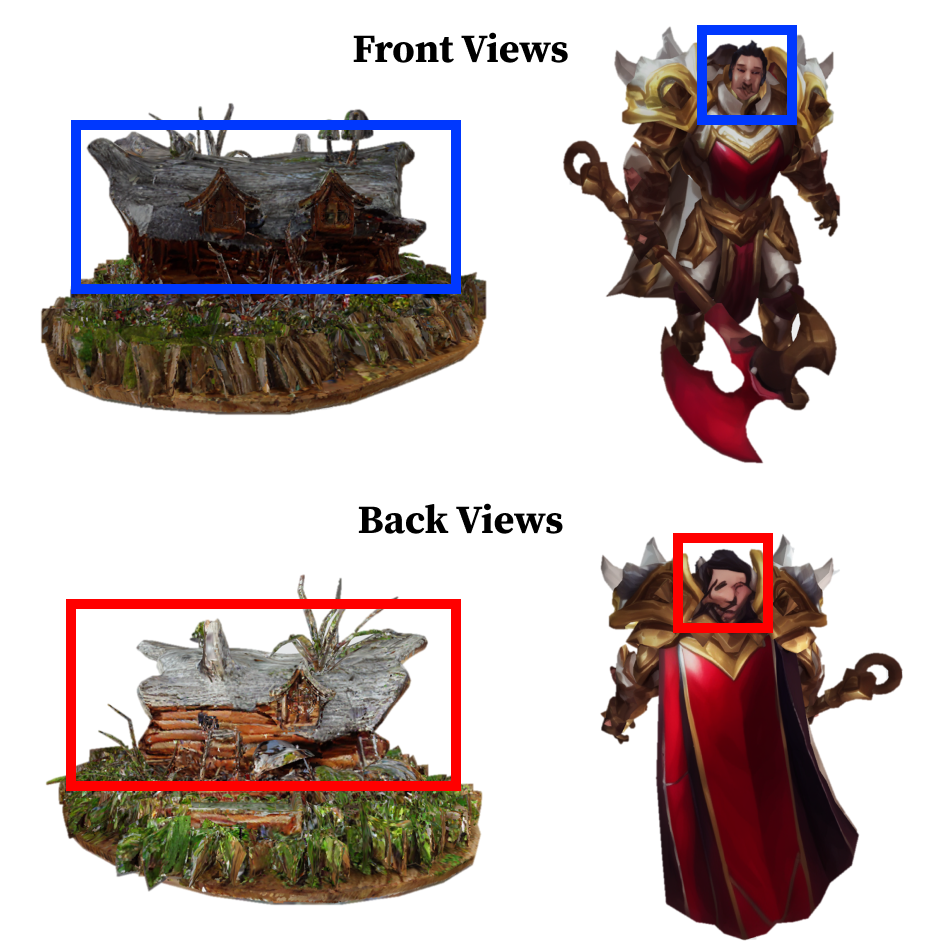}
        \caption{}
    \end{subfigure}
    \begin{subfigure}{0.32\textwidth}
        \centering
        \includegraphics[width=1\textwidth]{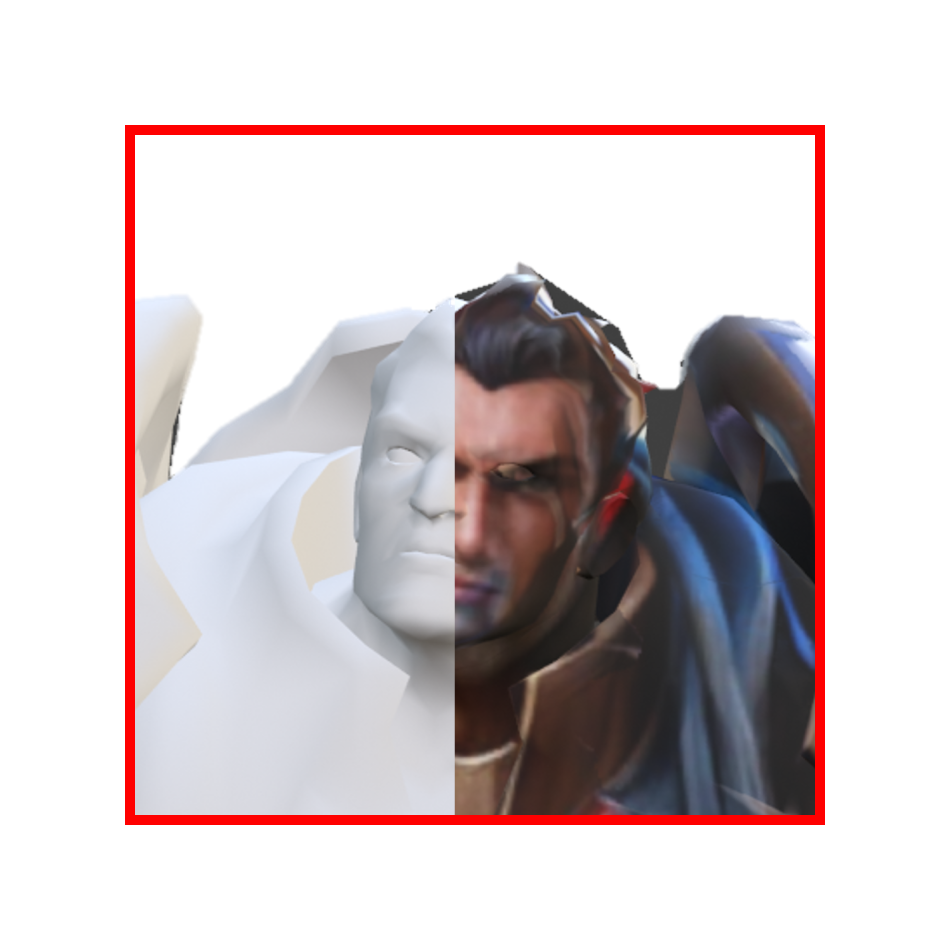}
        \caption{}
    \end{subfigure}
    \begin{subfigure}{0.32\textwidth}
        \centering
        \includegraphics[width=1\textwidth]{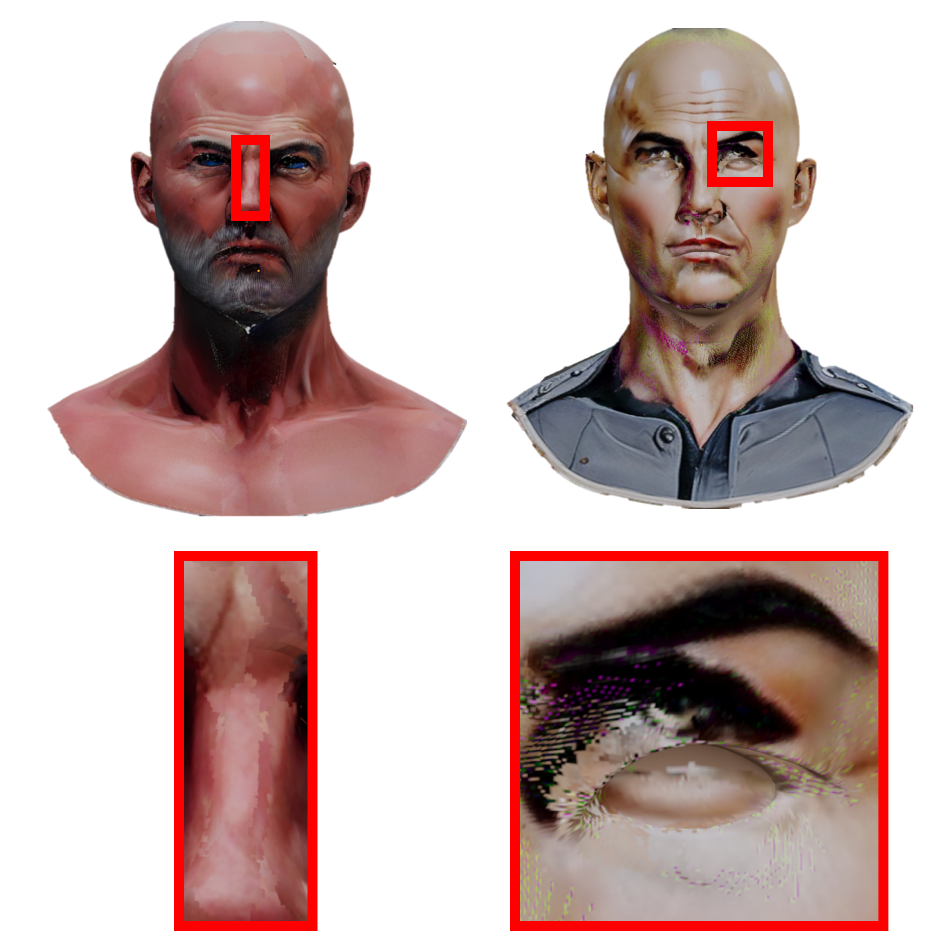}
        \caption{}
    \end{subfigure}
    \caption{2D diffusion-based texturing poses several challenges: (a) There exist inconsistencies between the front and back views; The textured character suffers from the Janus problem. (b) The texture is not aligned with the underlying mesh. (c) On the textured surfaces are many artifacts including seams; The left images are generated by Text2Tex, and the right by TEXTure.}
    \label{fig:problems}
\end{figure*}

In game, film productions and virtual/augmented reality industries, creating high-quality 3D assets is a time-consuming and resource-intensive process. In order to overcome such difficulties, researchers have recently developed \textit{generative models} for 3D content creation. 

Among the assets, \textit{textures} are particularly important because they significantly enhance the visual realism of the underlying 3D objects. A group of generative models trained on 3D datasets, such as Point-UV~\cite{point-UV}, generate textures by understanding the complete geometry of 3D objects. These models enable the creation of logical and occlusion-free textures. The dataset primarily used to train them is Objaverse~\cite{objaverse}. It contains over 800K 3D models of various categories and surpasses prior 3D datasets in size. However, it remains considerably smaller than the image datasets such as LAION~\cite{laion-400m, laion-5b}. This limited data availability prevents the models from generating a variety of textures.

TEXTure~\cite{TEXTure} and Text2Tex~\cite{text2tex} have for the first time proposed \textit{iterative} texture synthesis methods leveraging the prior knowledge of 2D \textit{diffusion models}~\cite{ddpm, ddim, ldm, dalle, dalle2}. Specifically, they use a pre-trained \textit{depth-to-image} diffusion model to capture the 3D object's geometric information. Expanding on this research, Paint3D~\cite{2023paint3} introduces a multi-view depth-aware texture sampling method to enhance view consistency and incorporates a position encoder on the UV space to remove lighting influences from the generated textures and also to inpaint the incomplete regions.

Leveraging the diversity and high expressiveness of image generation, the 2D diffusion-based approach has opened up the possibility of creating high-quality textures in a fast and easy way. However, the iterative nature of the approach encounters a problem due to the lack of comprehensive multi-view knowledge. Whereas humans generally evaluate an object from multiple angles, 2D diffusion model cannot, often resulting in inconsistent textures. 
Specifically, a single image is generated at a time in the iterations of TEXTure and Text2Tex, leading to the \textit{view-inconsistency} problem, as seen in the left images of Figure~\ref{fig:problems}-(a). To address this issue, Paint3D captures the object from a pair of symmetrical viewpoints at a time. As shown in the right images of Figure~\ref{fig:problems}-(a), however, this often causes the \textit{multi-face artifact}~\cite{dreamfusion}. Also called the \textit{Janus problem}, it occurs because the 2D diffusion model is not trained on multi-view datasets but is trained with a large number of frontal faces. 

On the other hand, previous 2D diffusion-based studies utilize the depth control only, and they adopt the Stable Diffusion 1.5 or 2.0 models~\cite{ldm}, which have a limited-size UNet architecture, allowing for control only with a $512\times512$ resolution image. This design choice leads to a lack of 3D awareness, often resulting in failure to align the texture with the underlying object, as illustrated in Figure~\ref{fig:problems}-(b). 

In each iteration of the diffusion-based methods, the generated image is projected back to the object's polygon mesh and then integrated into the evolving texture via UV mapping. This step usually produces unexpected artifacts including seams, as shown in Figure~\ref{fig:problems}-(c), where shown on the left are the results of Text2Tex and those of TEXTure are on the right.

This paper proposes a diffusion-based texture synthesis method designed to overcome the above-mentioned problems of the previous work, i.e., the proposed method addresses the challenges of synthesizing view-consistent, well-aligned, and seamless textures. It is dubbed RoCoTex for \textbf{Ro}bust method for \textbf{Co}nsistent \textbf{Tex}ture synthesis. RoCoTex employs a symmetrical view synthesis strategy, similar to Paint3D, and applies \textit{regional prompts} to the views to enhance view consistency. 
To generate textures that are aligned well with the underlying geometry, we leverage Stable Diffusion XL (SDXL)~\cite{sdxl}, which uses a three times larger UNet backbone than the previous versions of Stable Diffusion, and also multiple ControlNets for depths, normals and Canny edges, which help the network understand the underlying geometry. Finally, RoCoTex employs an efficient texture blending technique based on pixel confidences and a novel soft-inpainting technique based on Differential Diffusion~\cite{soft_inpainting} for reducing the seam regions.

The main contributions of this paper are summarized as follows:
\begin{enumerate}[left=0pt,topsep=0pt]
    \item We propose to combine a symmetrical view synthesis strategy with regional prompts, significantly enhancing view consistency and mitigating the Janus problem.
    \item We propose to combine SDXL with multiple ControlNets to generate well-aligned high-fidelity textures that capture structural features and intricate details.
    \item We introduce novel texture blending and soft-inpainting techniques, which reduce the seam regions successfully.
    \item Our extensive experiments demonstrate the robustness and consistency of RoCoTex, which outperforms the state-of-the-art methods.
\end{enumerate}

\section{Related Work}
\label{sec:related_work}
Texture generation techniques can be broadly categorized into two distinct approaches. 
The first approach involves leveraging a learning-based framework, which either directly learns from 3D datasets in the UV domain or utilizes score distillation sampling (SDS) loss to incorporate 2D diffusion priors~\cite{point-UV, chen2022auv, shap-e, chen2022tango, clipmesh, text2mesh, texturify, yeh2024texturedreamer, latent-nerf, fantasia3d}.
In contrast, the second approach entails generating 2D diffusion images conditioned on viewpoints, which are subsequently projected onto 3D meshes~\cite{TEXTure, text2tex, texro, 2023paint3, texfusion, zhang2024mapa, texture_field}.

\subsection{Diffusion Models}
Diffusion models~\cite{ldm, ddim, ddpm, sdxl} have become powerful tools for generative modeling, particularly in 2D image synthesis. These models learn data distributions by gradually adding and removing noise, enabling the creation of high-fidelity samples. Stable Diffusion~\cite{ldm} improves quality and stability by performing the diffusion process in a learned latent space rather than pixel space. Efforts to enhance 2D diffusion model's controllability include ControlNet~\cite{controlnet}, which incorporates additional input modalities such as semantic segmentation, depth, and edge maps to guide image generation. Stable Diffusion XL~\cite{sdxl} presents an extension of the Stable Diffusion framework, introducing an additional refinement stage and three times larger context dimensions.
\subsection{Learning-based Texture Generation}

\subsubsection{Learning from 3D Data}
AUV-Net ~\cite{chen2022auv} uses an autoencoder to generate UV maps from 3D mesh data, capturing geometric features and aligning textures to a canonical UV space. This method improves UV map quality and consistency but struggles with complex shapes. Point-UV~\cite{point-UV} introduces a UV diffusion model for 3D assets. It uses a coarse-to-fine pipeline, starting with a point diffusion model for low-frequency textures on the mesh surface, followed by a 2D diffusion model in the UV space to refine these textures. Learning from 3D data ensures consistent, mesh-aligned results, but it still faces challenges due to the limited datasets available.

\subsubsection{Learning from 2D Diffusion Prior}
SDS loss, as a 2D diffusion prior, is a loss function that guides the model to maintain the structural features of an image while transforming its style. It is primarily used by 3D generation models to synthesize 3D shapes and scenes from inputs like images and text~\cite{dreamfusion, latent-nerf, magic3d}. Dreamfusion~\cite{dreamfusion} introduces SDS loss for 3D generation, while Latent-Nerf~\cite{latent-nerf} applies SDS in the latent space. Expanding on these studies, TextureDreamer~\cite{yeh2024texturedreamer} employs a personalized diffusion model in conjunction with the PSGD (personalized geometric-aware score distillation) loss function to generate textures from input images. This method effectively transfers input textures onto the target mesh. Techniques utilizing 2D diffusion prior may be advantageous in terms of consistency, but they currently fall short of the desired fidelity.


\subsection{Texture Generation via 2D Image Projection}

\subsubsection{Recursive Sampling}
TexFusion~\cite{texfusion} proposed a sequential interlaced multi-view sampler that aggregates information from each viewpoint during every denoising step.
TexRO~\cite{texro} adopts an approach similar to TexFusion, with the distinction of performing the denoising process in the UV domain. Both methods demonstrate the ability to generate textured meshes with relatively high fidelity.
However, the recursive sampling process employed by these methods does not yet guarantee the same level of fidelity as 2D image generation.

\subsubsection{Iterative Texture Synthesis}

TEXTure~\cite{TEXTure} introduced an approach for texture generation leveraging 2D diffusion models.
Their method iteratively updates the texture by performing image-to-image translation using a 2D diffusion model, considering multiple viewpoints of the input mesh.
Text2Tex~\cite{text2tex} extended this approach by introducing an automatic view selection mechanism with a coarse-to-fine strategy.
Paint3D~\cite{2023paint3} further enhanced the process by utilizing a symmetric view inference process and a position encoder on the UV space for refinement.
While the symmetric view process was a motivating factor, there was a lack of explicit guidance in the generation stage.


\section{Proposed Method}
\label{sec:method}
\begin{figure}
    \centering
    \includegraphics[width=1.0\linewidth]{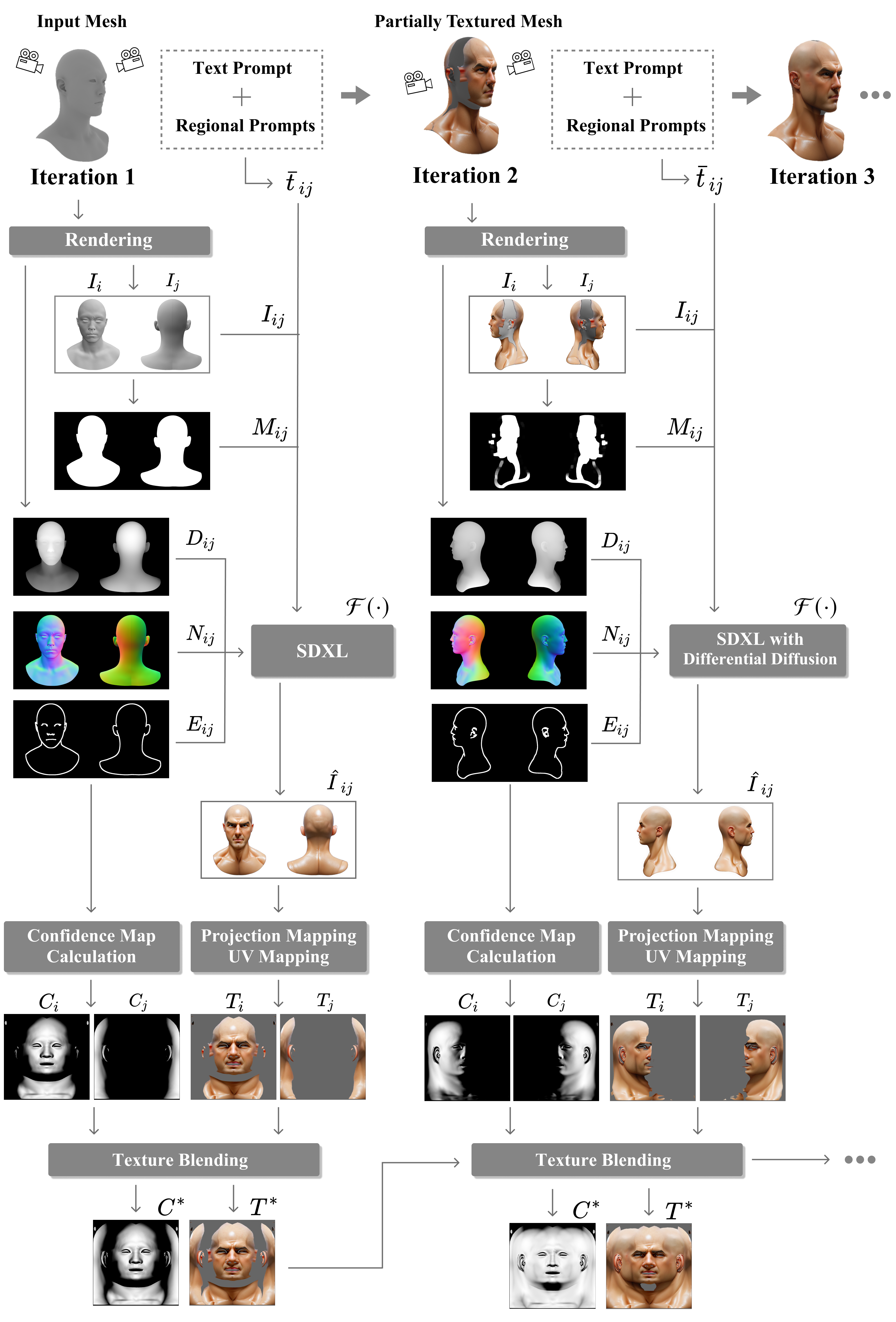}
    \caption{Overview of RoCoTex: The concatenated image ${I}_{ij}$, its inpainting mask ${M}_{ij}$, the depth map $D_{ij}$, the normal map $N_{ij}$, the edge map $E_{ij}$, and the SDXL output ${\hat{I}}_{ij}$ are of the same size, whereas the local confidence maps $C_i$ and $C_j$, the local textures $T_i$ and $T_j$, the global confidence map $C^*$ and the global texture $T^*$ are of the same size.     }
    \label{fig:overall_architecture}
\end{figure}

RoCoTex performs an iterative process so that the texture is progressively generated. Figure~\ref{fig:overall_architecture} illustrates the first two iterations made in RoCoTex, and their major steps are presented in the following subsections. 


\subsection{Symmetrical Views and Regional Prompts}

Generating a single image at a time often leads to \textit{context loss} and \textit{view inconsistency}~\cite{mvdream}.  
To address this problem, we generate two symmetrical views at a time, as Paint3D did. Figure~\ref{fig:overall_architecture} shows that the input mesh is rendered to generate two images, $I_i$ and $I_j$, which are then horizontally concatenated to define $I_{ij}$. 

Unfortunately, just taking such symmetrical views often causes a side effect: Because 2D diffusion models predominantly use front-view images~\cite{latent-nerf}, we may encounter the multi-face or Janus problem~\cite{magic3d, dreamfusion, mvdream}, as demonstrated in Figure~\ref{fig:problems}-(a). 

In order to tackle this challenge, we provide the \textit{regional prompts}, $t_i$ and $t_j$, for the symmetrical views. In the first iteration of Figure~\ref{fig:overall_architecture}, for example, $t_i$ is ``front view, (from front, front view focus)'' and $t_j$ is ``back view, (from back, back view focus)'' whereas the text prompt denoted as $t_0$ is ``Tom Cruise, bald, photorealistic.''
Using  the Regional Prompter~\cite{regional_prompt}, $\mathcal{R}(\cdot)$, the prompts are integrated: 
\begin{equation}
    \bar{t}_{ij} = \mathcal{R}(t_0, t_i, t_j).
    \label{eq:regional_prompt}
\end{equation}
By generating the symmetric views at a time, we can avoid context loss; furthermore, by providing the regional prompts for the views, we can mitigate the Janus problem.

\subsection{SDXL and Multiple ControlNets} 

For texture synthesis, preceding methods adopt Stable Diffusion as a backbone. 
However, the Stable Diffusion models trained on $512\times 512$ images have difficulties capturing high-fidelity details.
In order to generate high-quality textures with increased contextual understanding, we adopt Stable Diffusion XL (SDXL)~\cite{sdxl}, trained on 1K resolution with a three times larger UNet.

In our method,
SDXL takes not only the concatenated image, $I_{ij}$, but also its \textit{mask},  which specifies the ``untextured area'' of $I_{ij}$.
If we use the mask as is, however, artifacts can appear around the area's edges in the image generated by SDXL. To address this issue, we dilate the mask by 16 to 32 pixels, which we found to be optimal through experimentation. It is called an \textit{inpainting mask} and denoted as $M_{ij}$ in Figure~\ref{fig:overall_architecture}.

In order to make SDXL inference more 3D-aware, we leverage multiple ControlNets~\cite{controlnet} with $D_{ij}$, $N_{ij}$ and $E_{ij}$, which denote respectively the depth, normal and edge maps rendered from the input mesh. See Figure~\ref{fig:overall_architecture}. Our SDXL, denoted as $\mathcal{F}(\cdot)$, takes as input the concatenated image ${I}_{ij}$, its inpainting mask ${M}_{ij}$, the text and regional prompts $\bar{t}_{ij}$, and the conditions $\{D_{ij},N_{ij},E_{ij}\}$ to generate the image denoted as ${\hat{I}}_{ij}$: 
\begin{equation}
    {\hat{I}}_{ij} = \mathcal{F}({I}_{ij}, {M}_{ij}, \bar{t}_{ij}, D_{ij},N_{ij},E_{ij}; \tau_D, \tau_N, \tau_E),
    \label{eq:texturing}
\end{equation}
where $\tau_D$, $\tau_N$, and $\tau_E$ represent the ControlNets pre-trained on the depth, normal, and edge maps, respectively. 

Incorporating the complementary guidance allows us to generate textures that capture structural features and intricate details, improving the alignment and fidelity of the synthesized textures.

\subsection{Confidence-based Texture Blending}
\label{sec:conf}

\begin{figure}
    \centering
    \includegraphics[width=0.8\linewidth]{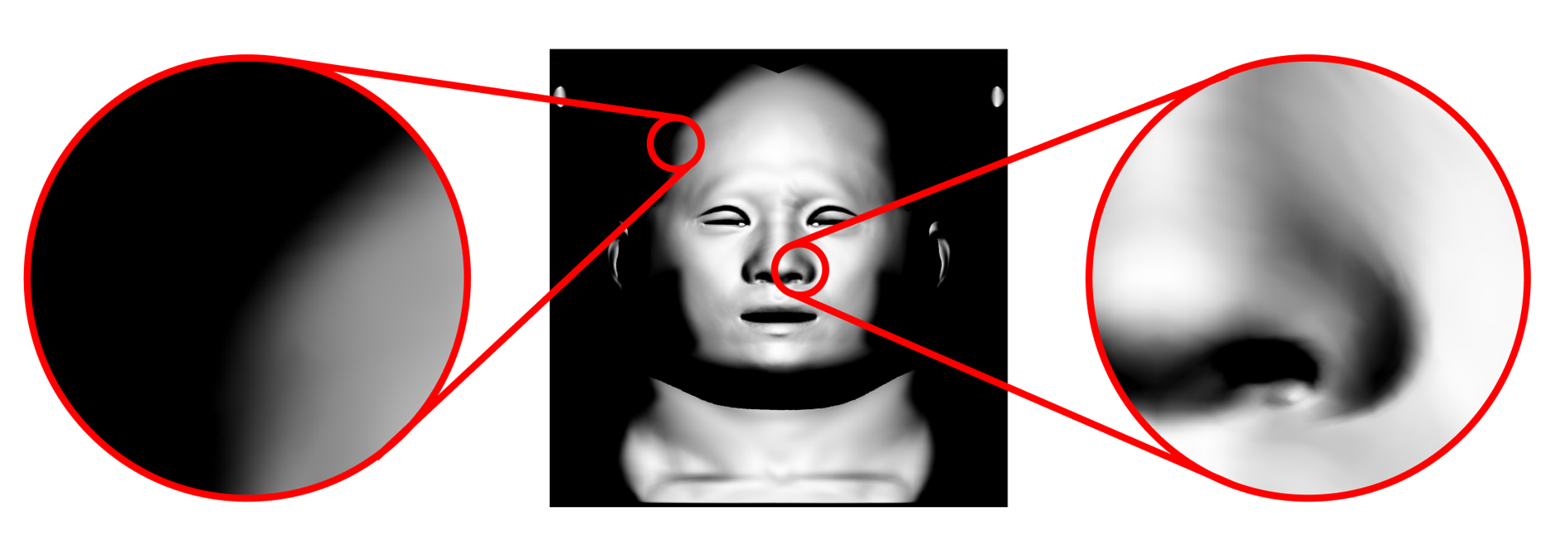}
    \caption{In the confidence map, $C_i$, the pixels located on the oblique triangles are given low confidences.}
    \label{fig:confidence-map-process}
\end{figure}

The image, ${\hat{I}}_{ij}$, generated by SDXL is decomposed into ${\hat{I}}_{i}$ and ${\hat{I}}_{j}$. Then, each image is back-projected onto the mesh by \textit{projection mapping} and then goes through \textit{UV mapping} to define a \textit{local} texture. Then, as shown in Figure~\ref{fig:overall_architecture}, the local textures, $T_i$ and $T_j$, are \textit{blended} into the \textit{global} texture, $T^*$, which evolves over iterations.  

Among the preceding methods, Text2Tex~\cite{text2tex} adopts a simple blending approach, which iteratively ``accumulates'' the local texture into the global one. It results in noticeable seams, as illustrated on the left of  Figure~\ref{fig:problems}-(c).  In contrast, TEXTure~\cite{TEXTure} performs an optimization, which directly updates the global texture every iteration. It successfully eliminates seams, but numerous unpredictable artifacts are generated, as shown on the right of  Figure~\ref{fig:problems}-(c).

To address these issues, we define the \textit{confidence} of each pixel in $T_i$ and $T_j$ and blend the pixel into $T^*$ using the confidence. 
Defined in the range $[0,1]$, the confidence is inversely proportional to the angle between the surface normal and the viewing direction. If a triangle is visible from the viewer but is angled obliquely, for example, its pixels are given smaller confidence values. The confidences are stored in the local confidence maps, $C_i$ and $C_j$, as shown in Figure~\ref{fig:overall_architecture}.  Figure~\ref{fig:confidence-map-process} shows the close-up views of confidence variation in $C_i$.  

$T^*$ and its global confidence map $C^*$ are updated as follows:
\begin{equation}
    T^* = \frac{T^* \cdot C^* + T_i \cdot C_i}{C^* + C_i + \epsilon},
    \label{eq:texture-blending}
\end{equation}
\begin{equation}
    C^* = C^* + C_i - C^* \cdot C_i,
    \label{eq:confidence}
\end{equation}
where $\cdot$ implies pixel-wise multiplications, and \(\epsilon\) is a small constant used for numerical stability. $T_j$ is blended into $T^*$ using $C_j$ in the same manner.

\subsection{Soft-inpainting with Differential Diffusion}
\begin{figure}[t]
    \centering
    \begin{subfigure}{0.4\columnwidth}
        \centering
        \includegraphics[width=\linewidth]
        {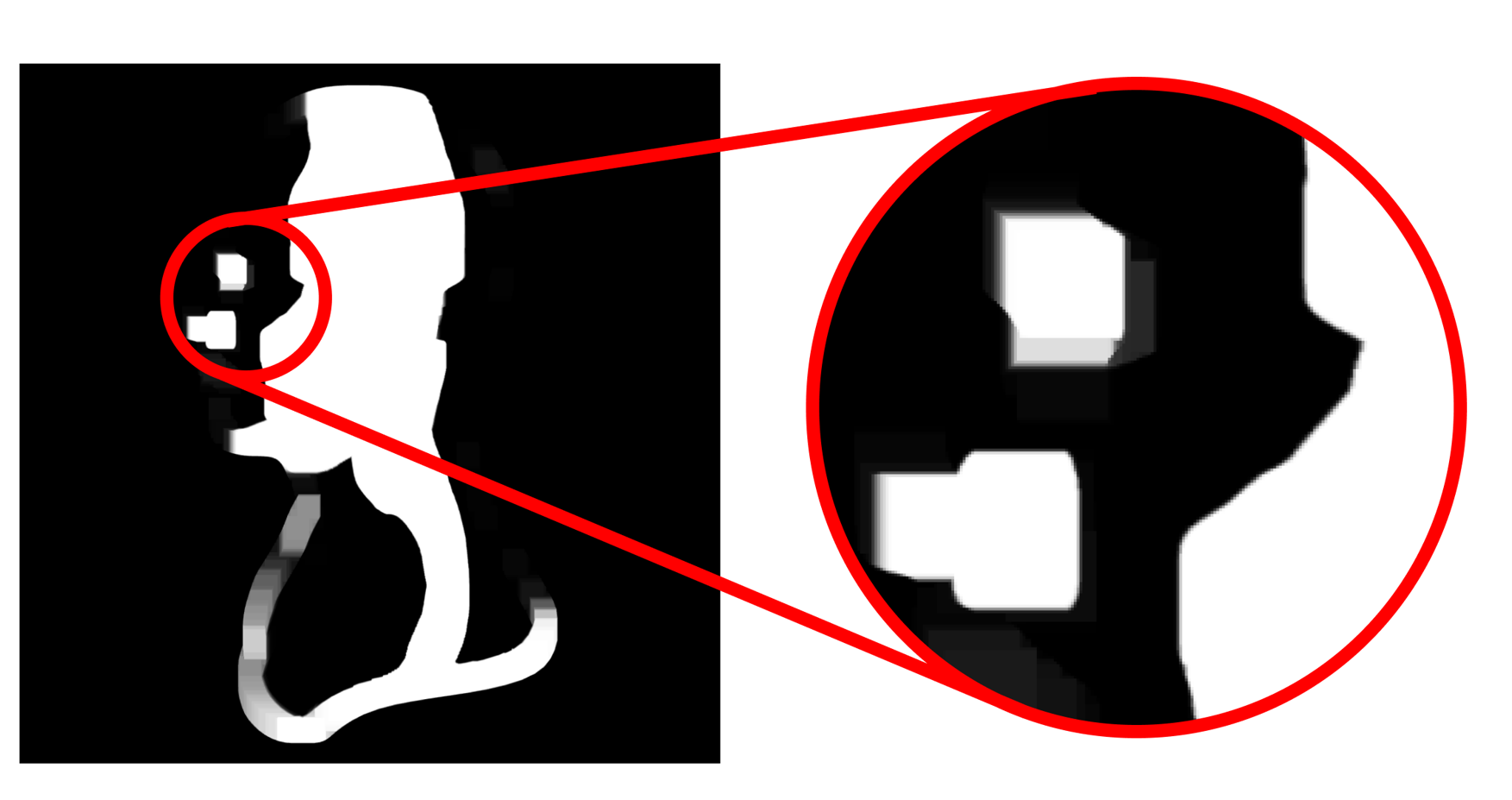}
        \caption{}
    \end{subfigure}
    \hspace{0.05\columnwidth}
    \begin{subfigure}{0.4\columnwidth}
        \centering
        \includegraphics[width=\linewidth]
        {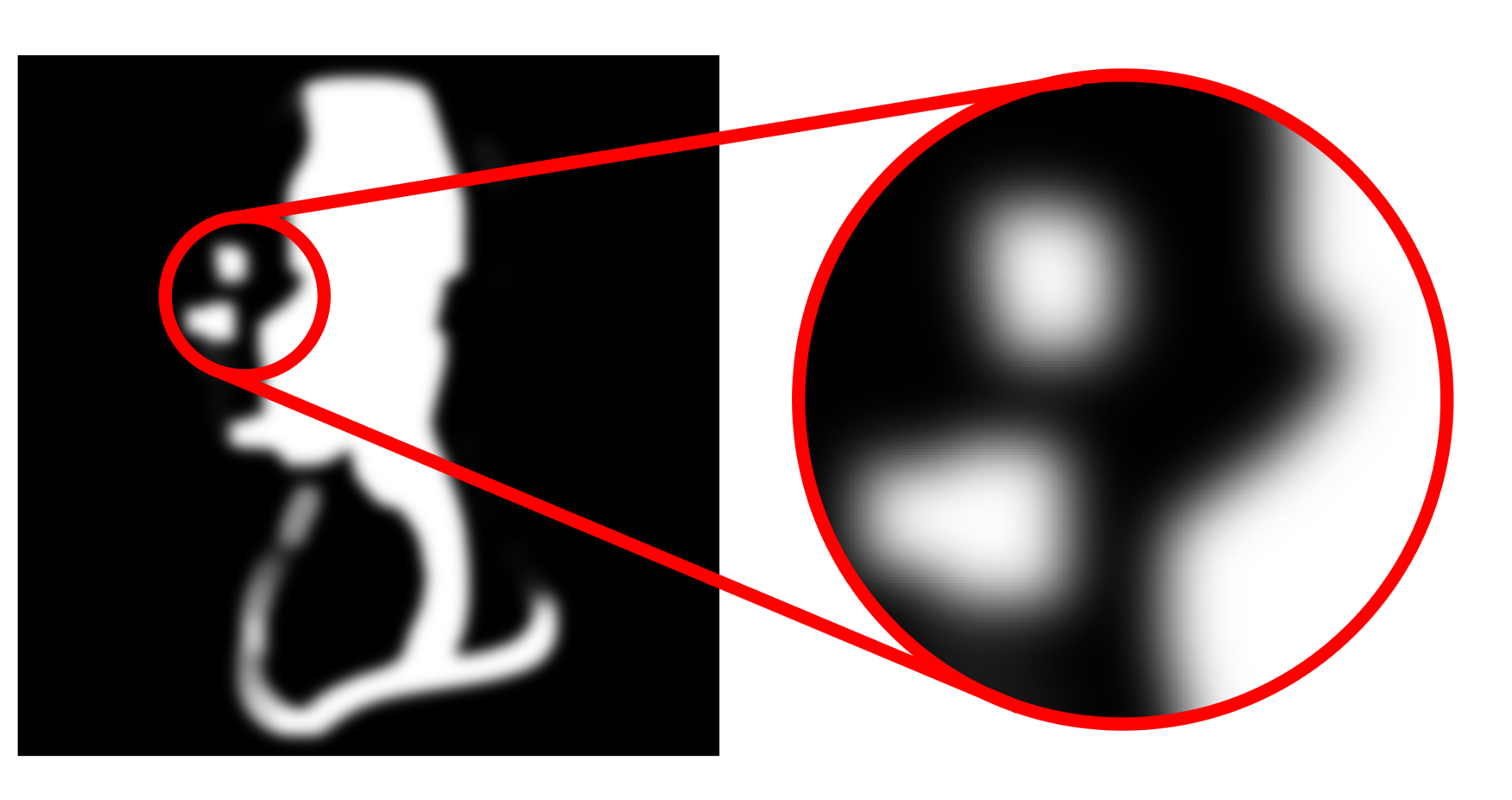}
        \caption{}
    \end{subfigure}
    \caption{Gaussian blurring of inpainting mask: (a) This shows the left part of $M_{ij}$, i.e., $M_i$. (b) $M_i$ is blurred. }
    \label{fig:blur-mask}
\end{figure}

In Figure~\ref{fig:overall_architecture}, consider the second iteration, where the partially textured mesh is taken as input. $I_{ij}$ and $M_{ij}$ are generated in the same manner as presented earlier. Now, the challenge is to \textit{inpaint} the ``untextured area'' specified by $M_{ij}$ while minimizing seams with the previously textured area. This challenge is called \textit{soft-inpainting}.

\begin{figure}[ht]
    \centering
    \begin{subfigure}{0.4\columnwidth}
        \centering
        \includegraphics[width=\linewidth]{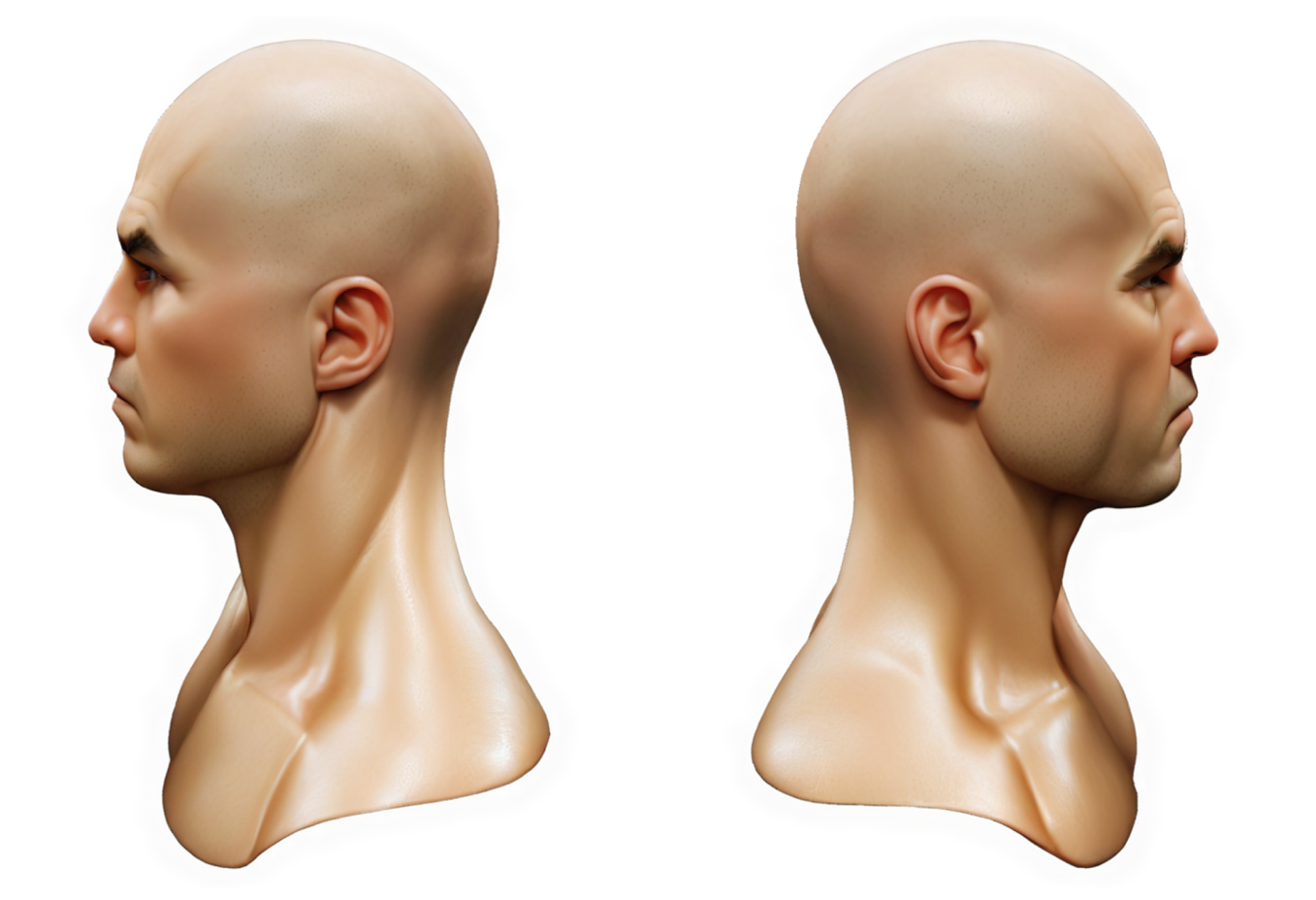}
        \caption{}
    \end{subfigure}
    \hspace{0.05\columnwidth}
    \begin{subfigure}{0.4\columnwidth}
        \centering
        \includegraphics[width=\linewidth]{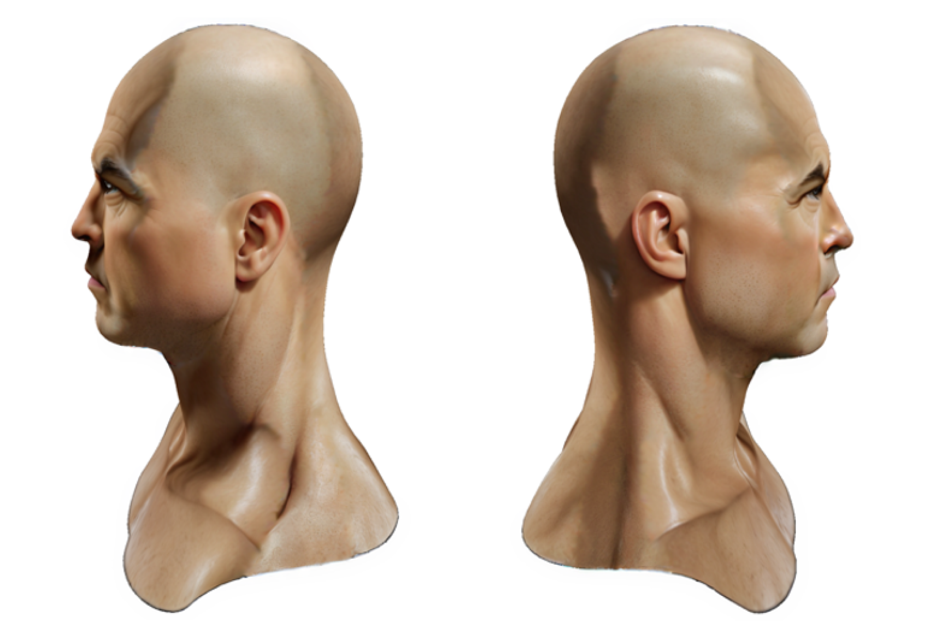}
        \caption{}
    \end{subfigure}
    \caption{Comparison of inpainting: (a) In RoCoTex, \textit{continuous} denoising strengths are used for \textit{soft-inpainting}. (b) In Text2Tex, a \textit{constant} denoising strength is assigned to a region of the generation mask, producing noticeable seams.}
    \label{fig:soft_inpainting}
\end{figure}

For soft-inpainting, (1) $M_{ij}$ is Gaussian blurred, as shown in Figure~\ref{fig:blur-mask}, and (2) SDXL is integrated with an advanced diffusion technique named \textit{Differential Diffusion}~\cite{soft_inpainting}, which allows to give each pixel its own strength. (For details on Differential Diffusion, readers are referred to the original paper authored by Levin and Fried~\cite{soft_inpainting}.)
By inputting the blurred $M_{ij}$ into the SDXL integrated with Differential Diffusion, it becomes possible to compute \textit{continuous} denoising strengths using the \textit{continuous} values of the blurred $M_{ij}$. Figure~\ref{fig:soft_inpainting}-(a) shows the close-up view of ${\hat{I}}_{ij}$, which is generated via soft-inpainting.
Observe that the ``untextured area'' has been inpainted with  little seams with the previously textured area.

\subsection{Discussion}
In the same way as presented in Section~\ref{sec:conf}, Text2Tex~\cite{text2tex} also computes the ``confidences.''
However, the usage of confidences in Text2Tex is different from ours.  
In Text2Tex, the confidences are used to create the so-called generation mask, which is composed of ``new'' (with the denoising strength $\gamma=1$), ``update'' ($\gamma=0.5$) and ``keep'' ($\gamma=0$) regions. 
This mask is used for the denoising steps within the Stable Diffusion model. 
Unfortunately, such an attempt to reduce seams is not effective because they use a constant denoising strength for each region. 
Figure~\ref{fig:soft_inpainting}-(b)
shows the inpainting result, which reveals noticeable seams. TEXTure~\cite{TEXTure} employs a similar technique, suffering from the same problem.


\section{Experiments}
\label{sec:experiments}


\begin{figure*}[t]
    \centering
    \begin{subfigure}{0.49\textwidth}
        \centering
        \includegraphics[width=1\textwidth]{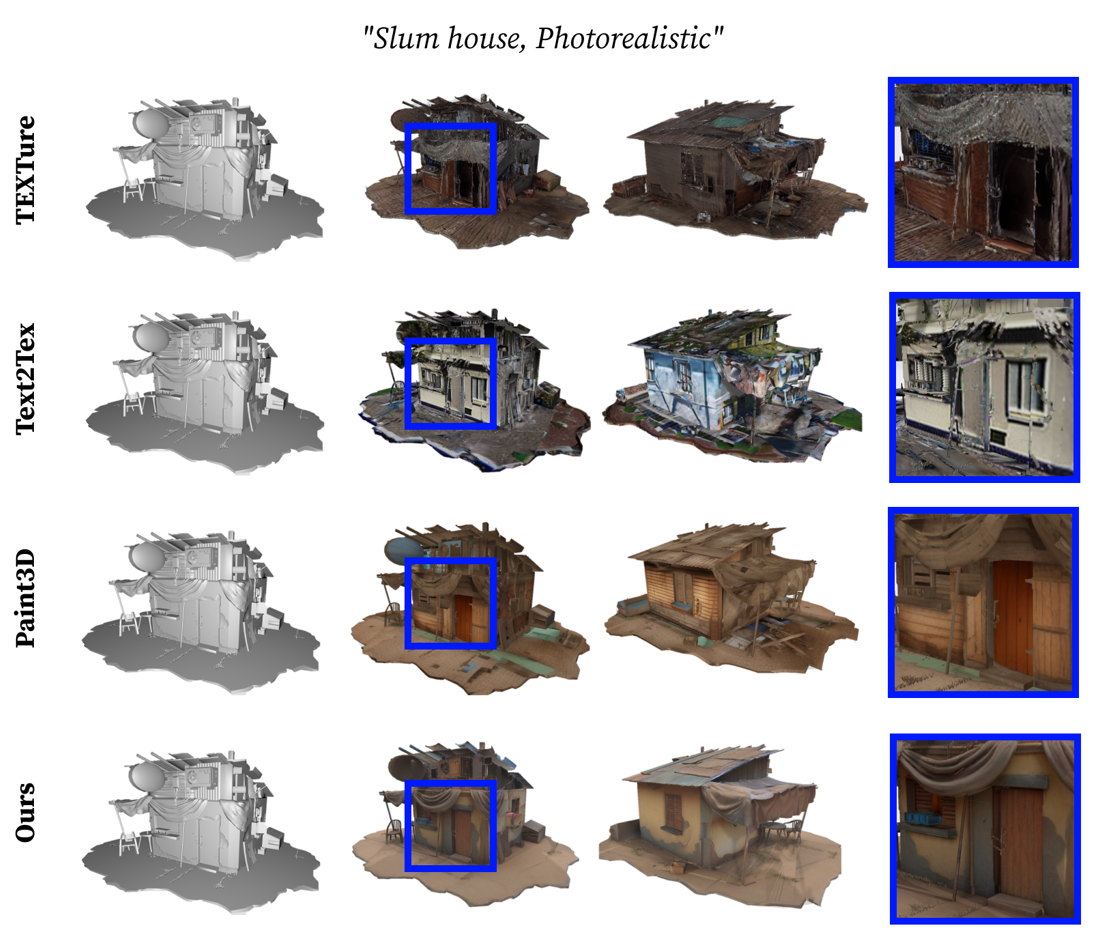}
        \caption{}
    \end{subfigure}
    \begin{subfigure}{0.42\textwidth}
        \centering
        \includegraphics[width=1\textwidth]{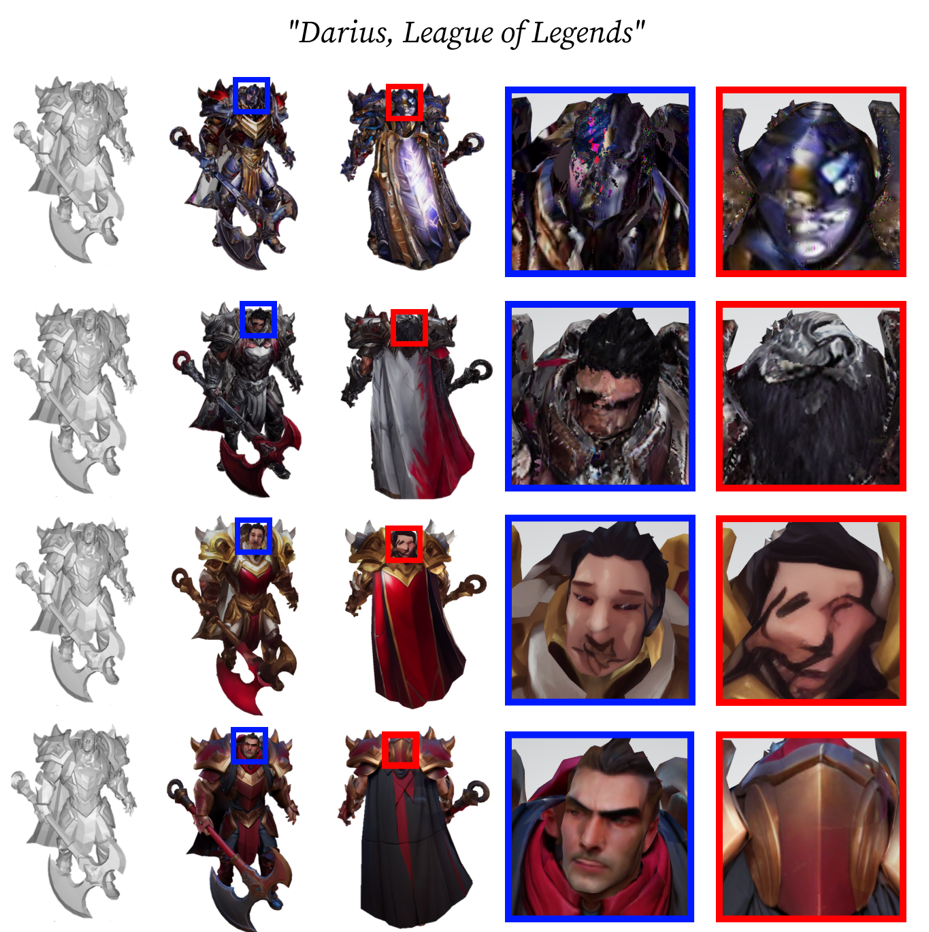}
        \caption{}
    \end{subfigure}
    \caption{Qualitative comparisons.}
    \label{fig:comparison}
\end{figure*}

The robustness and consistency of RoCoTex are validated through various experiments. As the state-of-the-art baseline methods, we use TEXTure~\cite{TEXTure}, Text2Tex~\cite{text2tex}, and Paint3D~\cite{2023paint3}, the source codes of which are publicly available.

\subsection{Implementation Details}

For the inference process, we employ SDXL~\cite{sdxl} as our generation backbone, at the custom resolution of $2048 \times 1024$ due to  symmetrical view synthesis. The depth, normal and Canny edge ControlNet weights are set to 0.5.

The 3D models are taken either from Objaverse~\cite{objaverse} or from the game developing studio in which a subset of this paper's authors work.
We use Trimesh for handling triangle meshes and Pyrender for rendering. 
All experiments are conducted on an NVIDIA A100 GPU.

\subsection{Qualitative Results}

Using an asset named ``Slum house,'' 
Figure~\ref{fig:comparison}-(a) compares qualitatively three baselines and our method. The first 
column shows the untextured mesh, and the second and third columns are the front and back views of the textured mesh, respectively. Especially in Text2Tex, the view-inconsistency problem is clearly visible. On the other hand, considering the location of the front door at the first column, it can be easily observed that both TEXTure and Text2Tex suffer from the problem of misaligned textures. In contrast, the doors are relatively well aligned in Paint3D and RoCoTex. In Paint3D, however, the drapes above the door reveal artifacts.  

In Figure~\ref{fig:comparison}-(b), the head of ``Darius'' in the front view is zoomed-in in blue boxes and that in the back view is in red boxes. TEXTure and Text2Tex suffer from many artifacts in both views. Even though Paint3D captures a pair of symmetrical views at a time, it suffers from the Janus problem, which is resolved using the regional prompts in RoCoTex.

Figure~\ref{fig:comparison} shows that our method demonstrates a more comprehensive understanding of both the prompts and the underlying geometry, generating high-quality well-aligned textures in general. 

\subsection{Quantitative Results}


The generated textures are evaluated using Kernel Inception Distance (KID)~\cite{kid}, which is a commonly used image quality and diversity metric for  generative models. 
Table~\ref{tab:user_study} shows that RoCoTex achieves the lowest KID score, indicating higher quality and diversity of the generated images. 

\subsection{User Study}
\begin{table}
\vspace{0.5cm}
\setlength{\tabcolsep}{3pt}
\caption{Quantitative results for the KID and user study.}
\centering
\begin{tabular}{ccccc}
\toprule
\multirow{2}{*}{Method} & \multirow{2}{*}{KID $\downarrow$} & \multicolumn{3}{c}{User Study (\%)} \\
\cmidrule(lr){3-5}
  & & Quality $\uparrow$ & Consistency $\uparrow$ & Alignment $\uparrow$ \\
\midrule
TEXTure & 10.34 & 3.0 & 1.8 & 2.6  \\
\cmidrule(lr){1-1} \cmidrule(lr){2-2} \cmidrule(lr){3-5}
Text2Tex & 8.15 & 10.5 & 7.3 & 12.7  \\
\cmidrule(lr){1-1} \cmidrule(lr){2-2} \cmidrule(lr){3-5}
Paint3D & 6.98 & 9.5 & 15.0 & 12.7  \\
\midrule
\textbf{RoCoTex} & \textbf{4.03} & \textbf{77.0} & \textbf{76.0} & \textbf{71.9} \\
\bottomrule
\end{tabular}
\label{tab:user_study}
\end{table}

Table~\ref{tab:user_study} also shows the results of user study, which is made in terms of quality, consistency and alignment. 
Involved in the user study are 40 participants with varying levels of expertise in 3D modeling and texturing.
Each participant reviews 10 different assets, using a 3D web viewer. (Watch the video.)
For each asset, the participants are instructed to identify the method that most effectively demonstrates the texture's quality, consistency and alignment with the underlying object. 
To mitigate potential bias, the sequence of presenting four methods is randomized.
The results, detailed in Table~\ref{tab:user_study}, demonstrate that RoCoTex outperforms the other methods across all criteria. 

\subsection{Ablation Study}

With RoCoTex, the ablation study is made for proving the effects of symmetrical view synthesis, regional prompts, multiple ControlNets, confidence maps and soft-inpainting.

\subsubsection{Symmetrical View Synthesis}
\begin{figure}[t]
    \centering
    \begin{subfigure}{0.49\columnwidth}
        \centering
        \includegraphics[width=\linewidth]{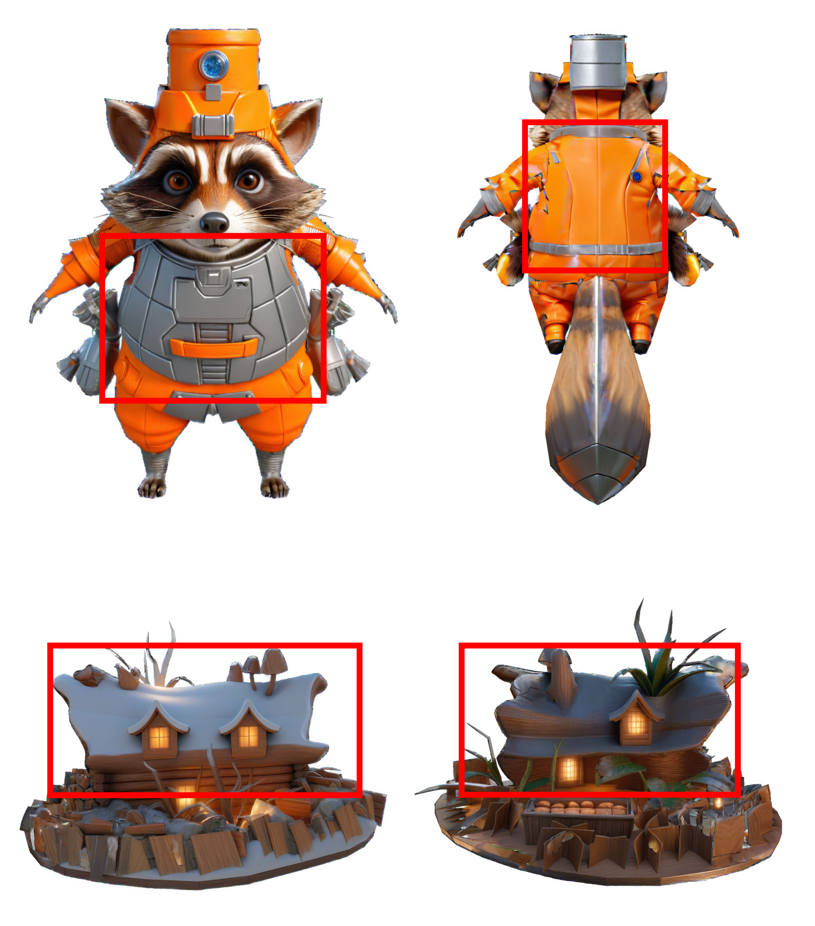}
        \caption{}
    \end{subfigure}
    \begin{subfigure}{0.49\columnwidth}
        \centering
        \includegraphics[width=\linewidth]{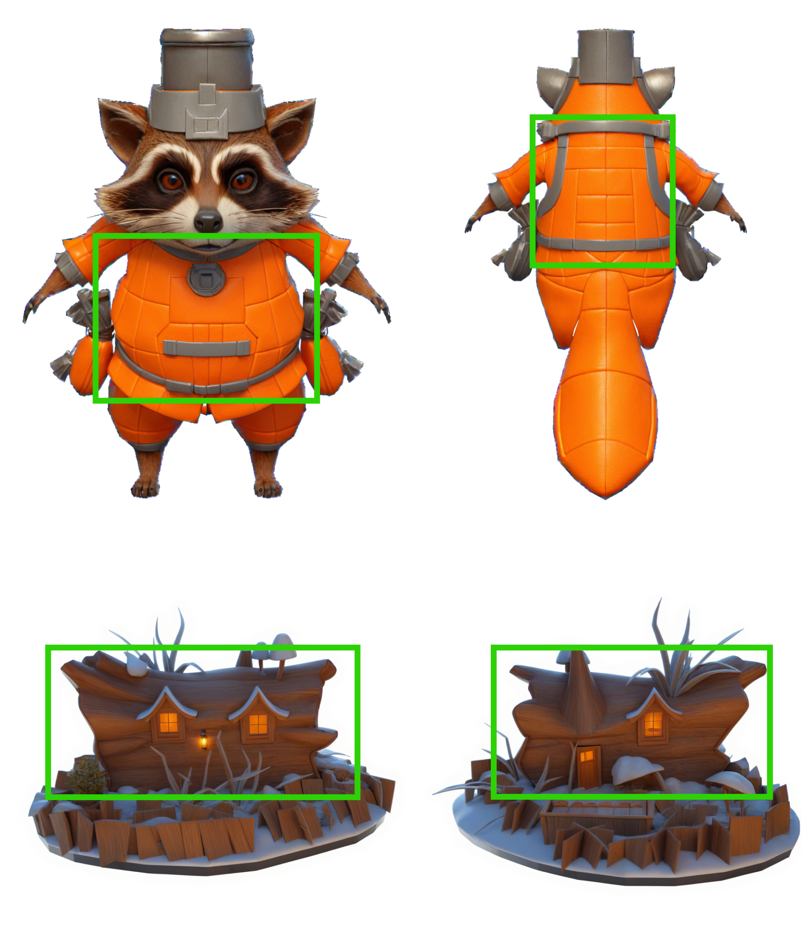}
        \caption{}
    \end{subfigure}
    \caption{Symmetrical view synthesis: (a) The 3D objects are textured by generating a single image at a time. (b) The objects are textured by generating a pair of symmetrical views at a time. They are consistently textured.}
    \label{fig:cycle-ablation}
\end{figure}

With two assets, Figure~\ref{fig:cycle-ablation}-(a) shows the front and back views obtained by generating a single image at a time, and Figure~\ref{fig:cycle-ablation}-(b) shows the same views obtained via symmetrical view synthesis.  It can be clearly observed that the symmetrical view approach improves the consistency of textures in 3D models.

\subsubsection{Regional Prompts}
\begin{figure}[ht]
    \centering
    \begin{subfigure}{0.49\columnwidth}
        \centering
        \includegraphics[width=\linewidth]{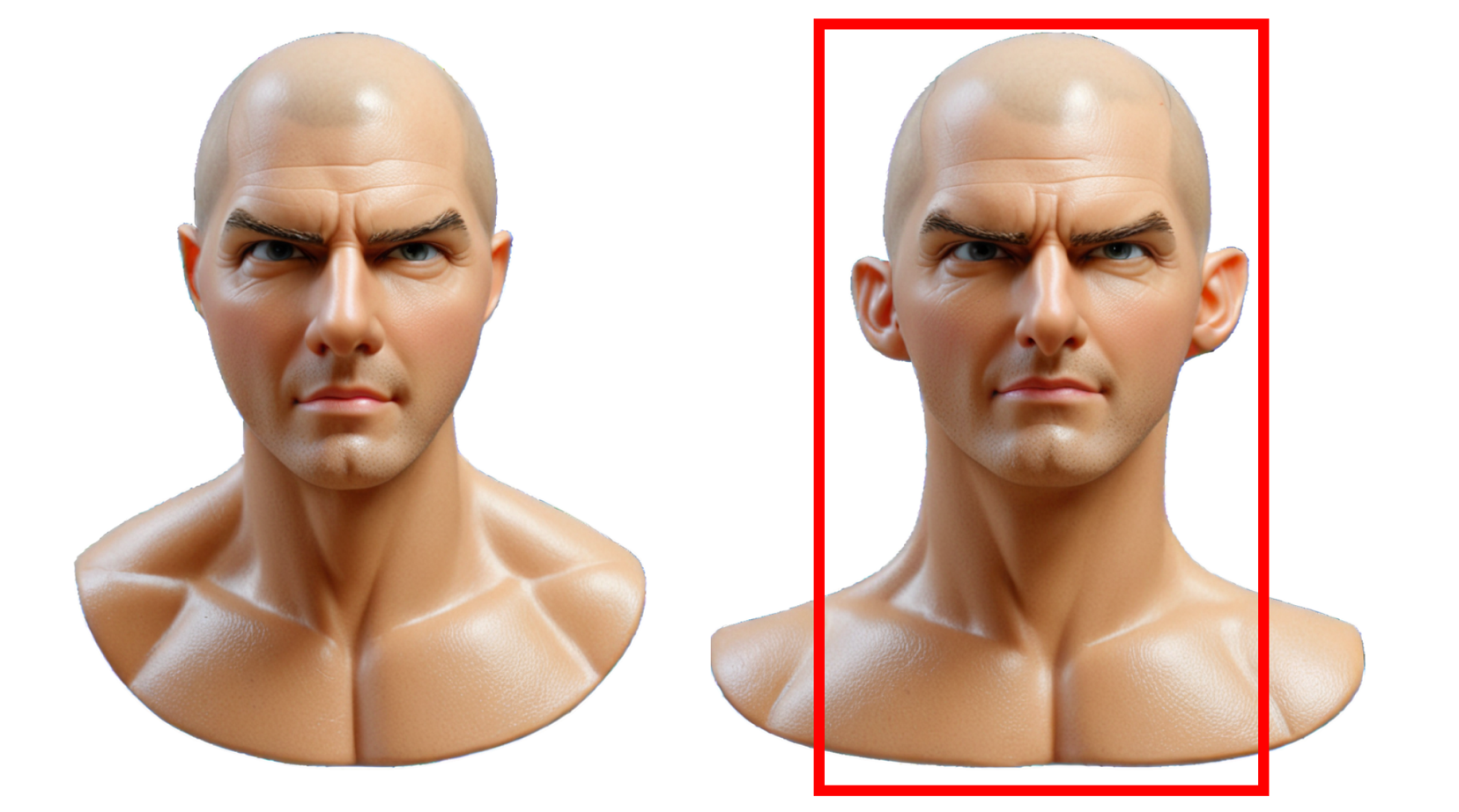}
        \caption{}
    \end{subfigure}
    \hfill
    \begin{subfigure}{0.49\columnwidth}
        \centering
        \includegraphics[width=\linewidth]{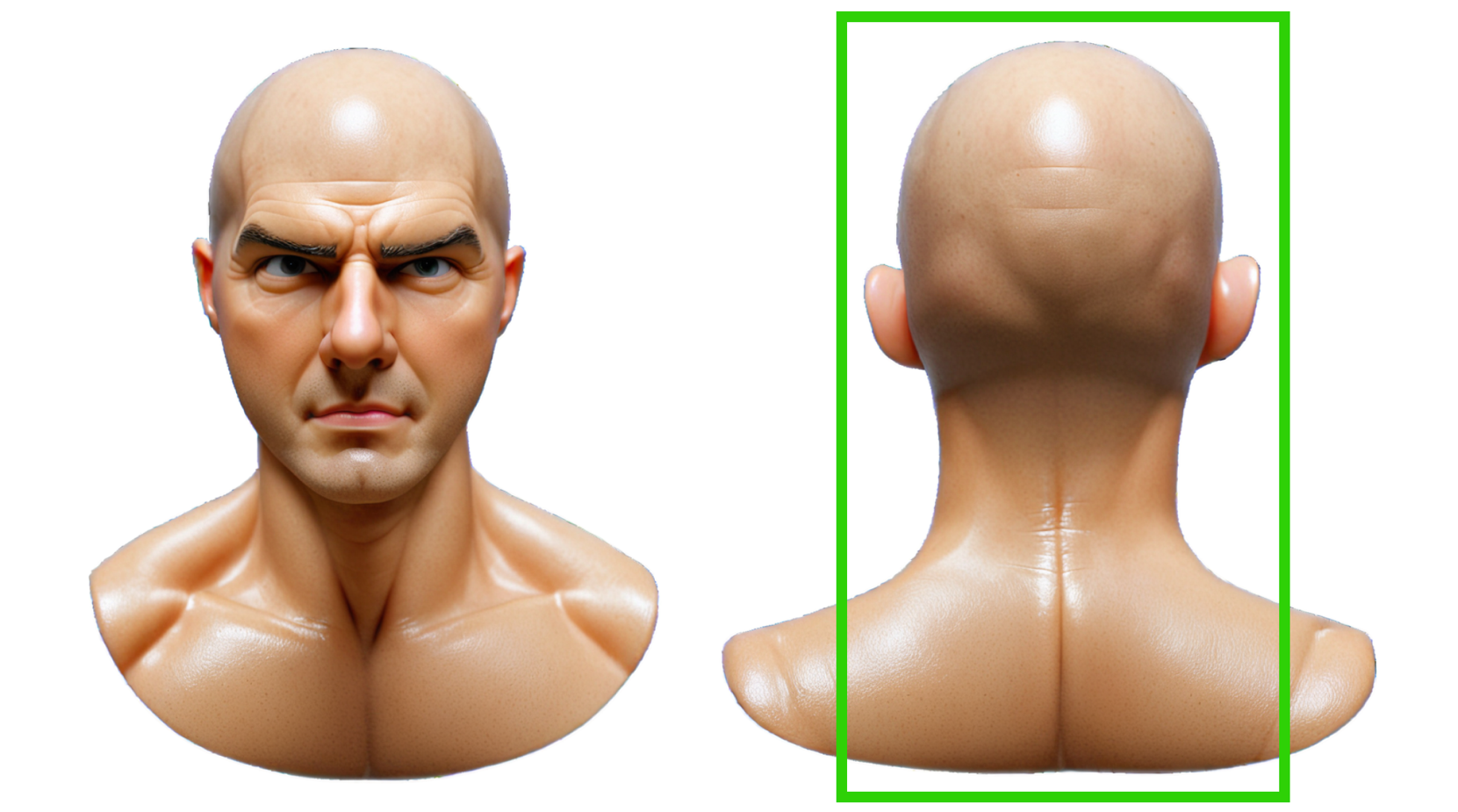}
        \caption{}
    \end{subfigure}
    \caption{Regional prompts: (a) On the left is the front view; On the right is the back view with the Janus problem. (b) The regional prompts resolve the problem.}
    \label{fig:regional-prompt}
\end{figure}


Figure~\ref{fig:regional-prompt} compares two texturing results: (a) without the regional prompts, and (b) with the regional prompts. Observe that our method might suffer from the Janus problem without the appropriate regional prompts. 

\subsubsection{Multiple ControlNets}

\begin{figure}[ht]
    \centering
    \begin{subfigure}{0.48\columnwidth}
        \centering
        \includegraphics[width=\linewidth]{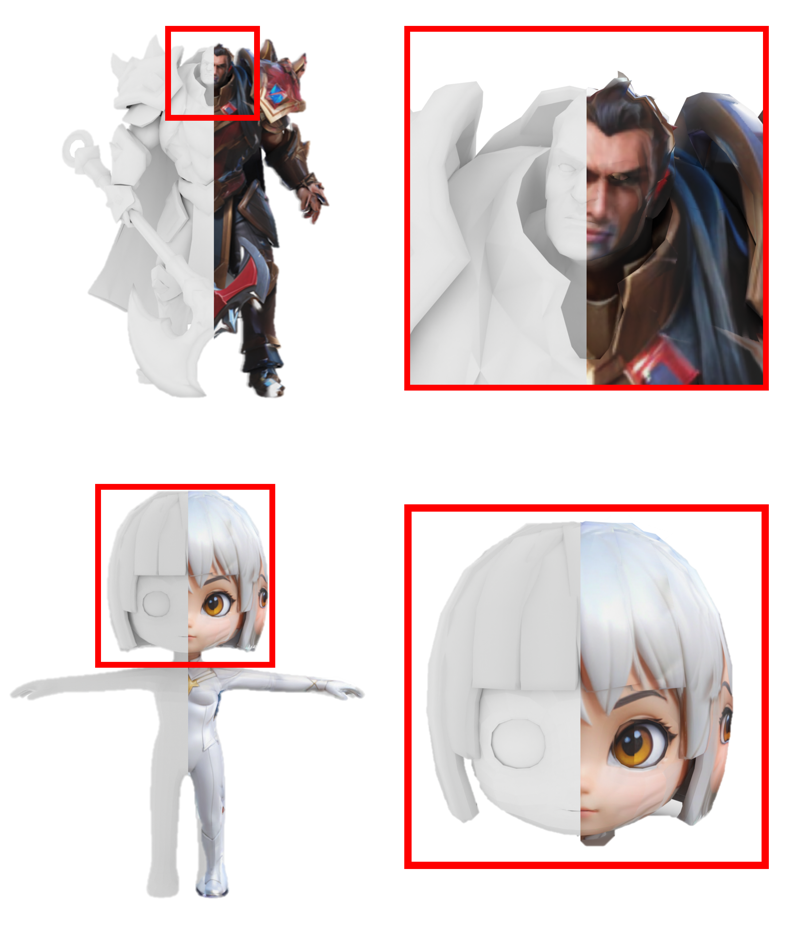}
        \caption{}
    \end{subfigure}
    \hfill
    \begin{subfigure}{0.48\columnwidth}
        \centering
        \includegraphics[width=\linewidth]{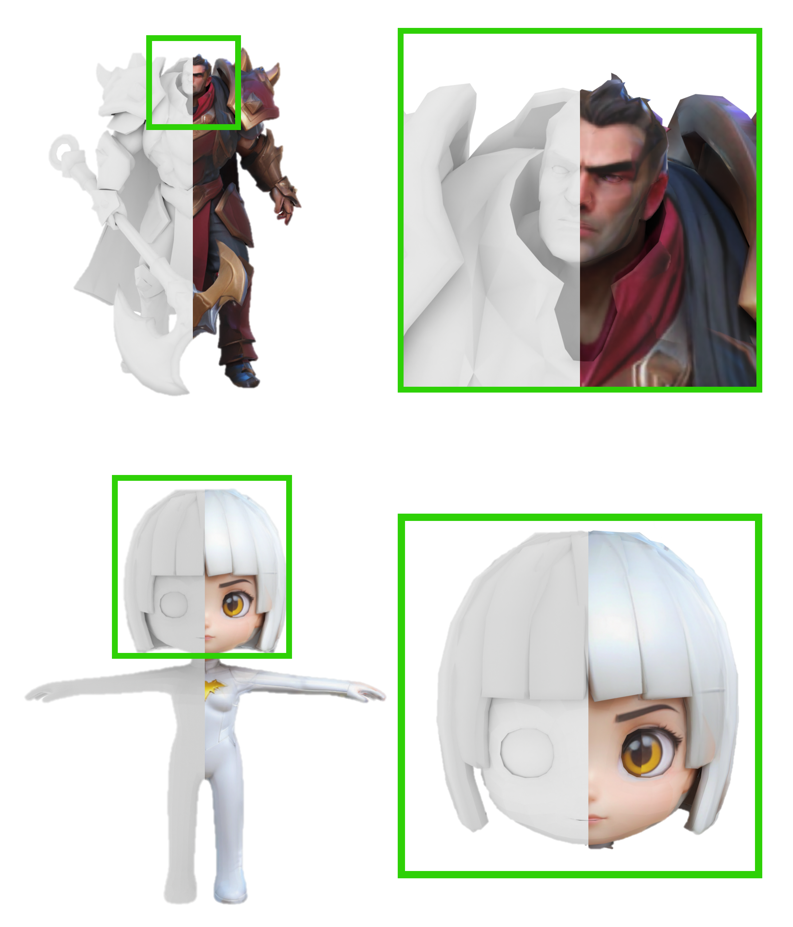}
        \caption{}
    \end{subfigure}
    \caption{Multiple ControlNets: (a) Depth control only. (b) Multiple controls.}
    \label{fig:controlnet-ablation}
\end{figure}

Figure~\ref{fig:controlnet-ablation} compares the texturing results: (a) with the depth control only, and (b) with multiple controls, i.e., with depth, normal and edge guidance. In ``Darius,'' the misalignment problem is resolved using multiple ControlNets. In ``White cute hero,'' the hair style appears distorted when we use only the depth control. (Note the difference from the untextured mesh on the left.) In contrast, multiple ControlNets bring about correct results.

\subsubsection{Confidence Maps}
\begin{figure}[ht]
    \centering
    \begin{subfigure}{0.49\columnwidth}
        \centering
        \includegraphics[width=\linewidth]{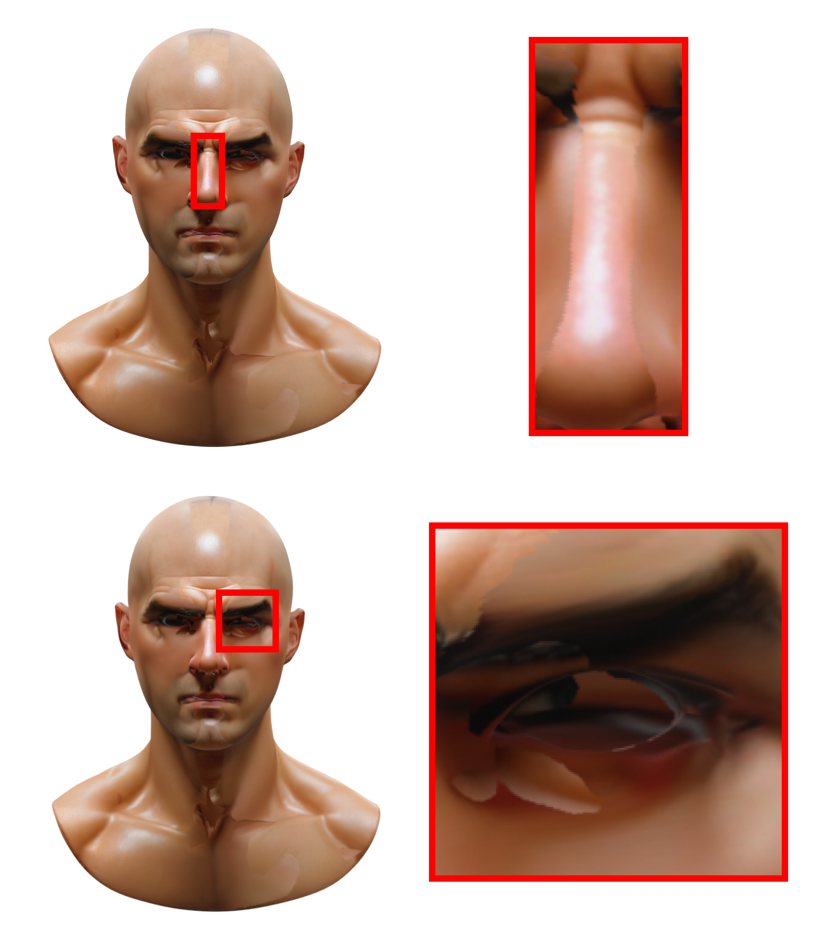}
        \caption{}
    \end{subfigure}
    \hfill
    \begin{subfigure}{0.49\columnwidth}
        \centering
        \includegraphics[width=\linewidth]{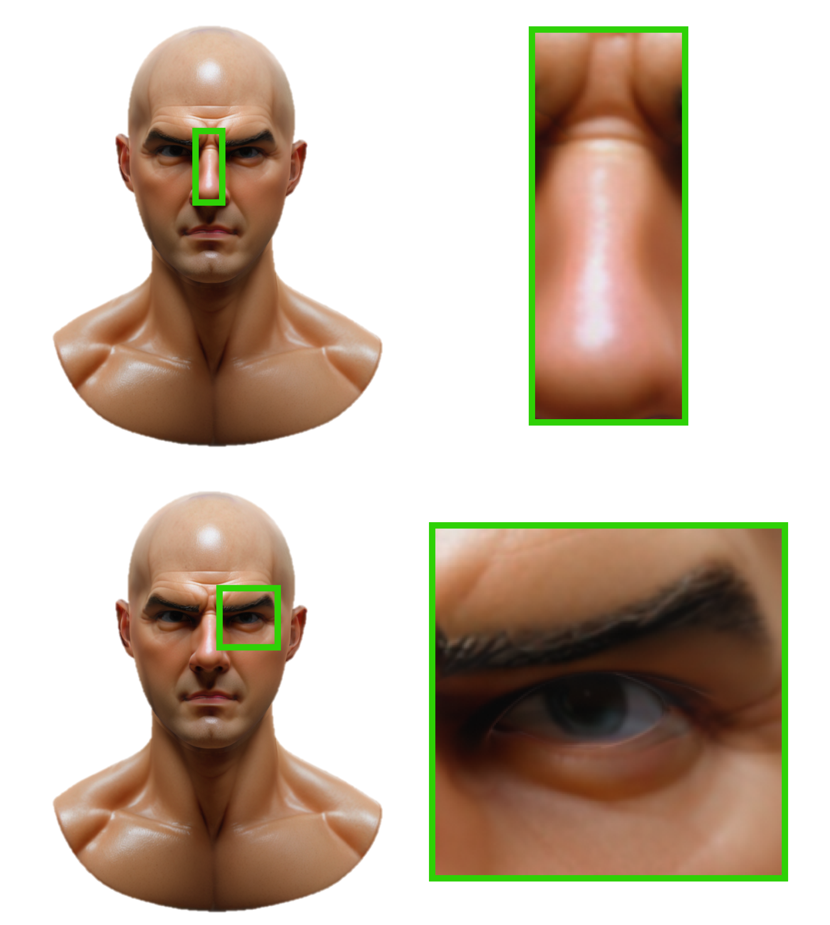}
        \caption{}
    \end{subfigure}
    \caption{Confidence maps: (a) Simple accumulation. (b) Blending using confidence maps.}
    \label{fig:blending-ablation}
\end{figure}

Figure~\ref{fig:blending-ablation}-(a) shows the result of iteratively ``accumulating'' the local textures onto the global texture. (The blending methods of Text2Tex and Paint3D are emulated in RoCoTex.) In contrast, Figure~\ref{fig:blending-ablation}-(b) shows the result of our confidence-based texture blending, which produces superior results. 

\subsubsection{Soft-inpainting}
\begin{figure}[ht]
    \centering
    \begin{subfigure}{0.49\columnwidth}
        \centering
        \includegraphics[width=\linewidth]{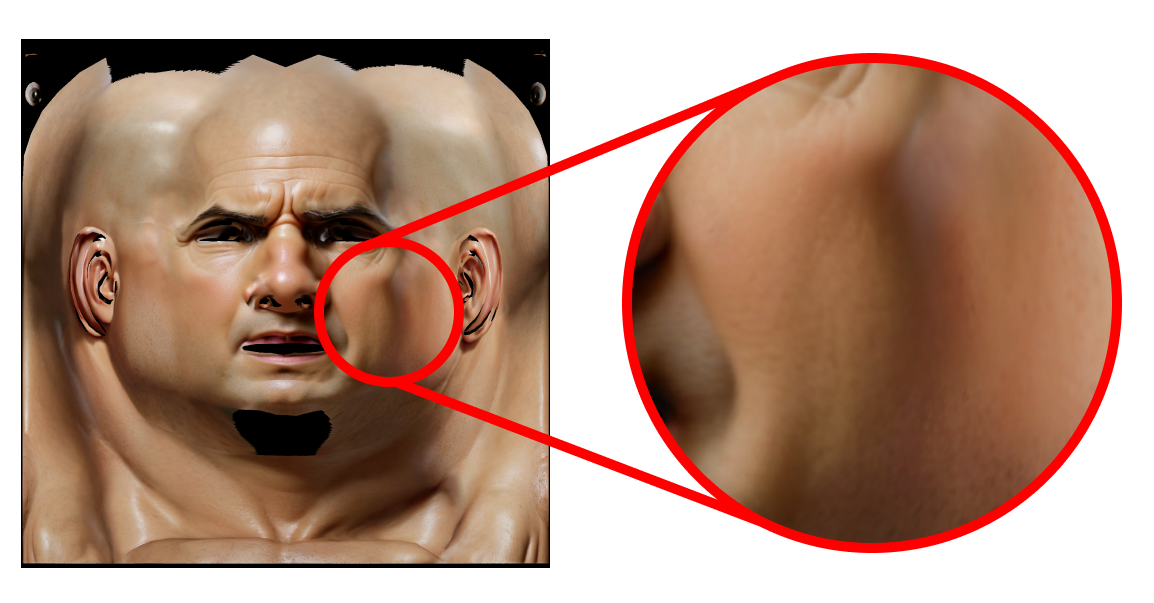}
        \caption{}
    \end{subfigure}
    \hfill
    \begin{subfigure}{0.49\columnwidth}
        \centering
        \includegraphics[width=\linewidth]{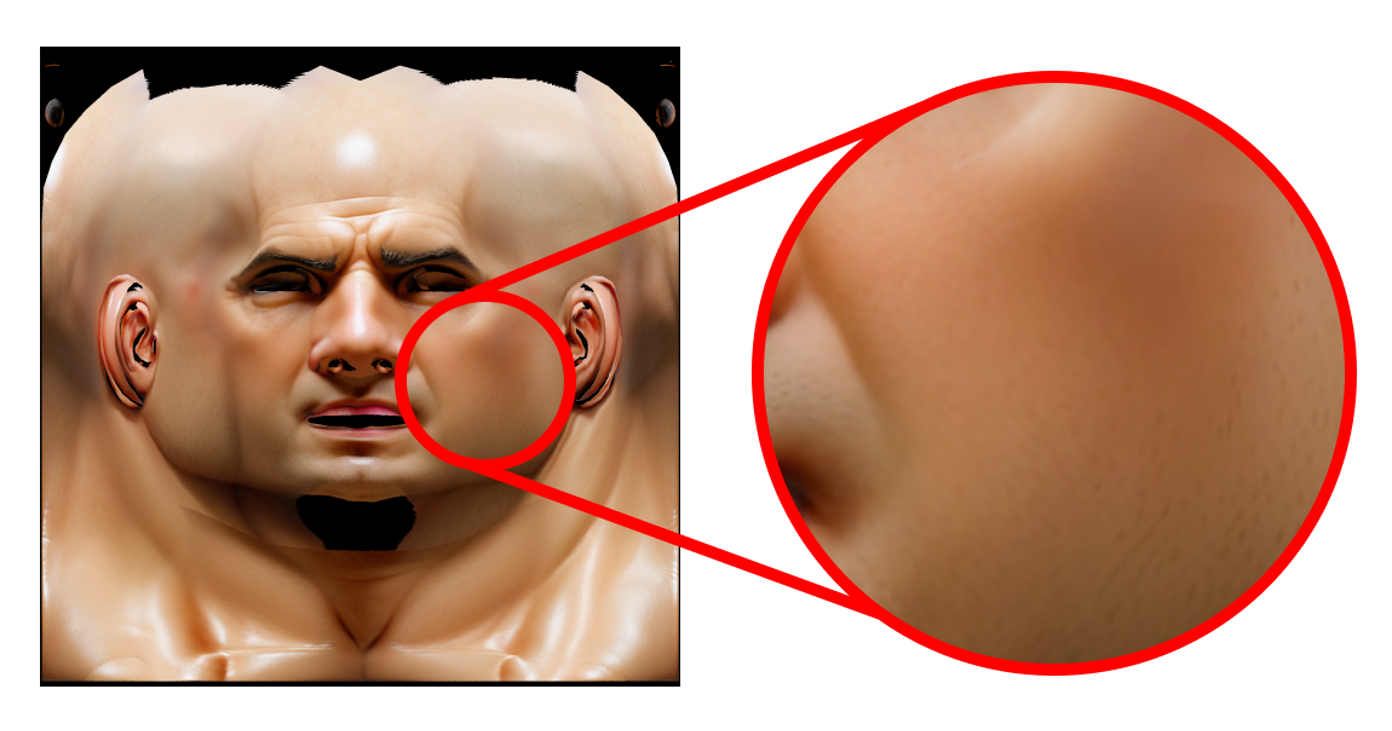}
        \caption{}
    \end{subfigure}
    \caption{Soft-inpainting: (a) Without soft-inpainting. (a) With soft-inpainting.}
    \label{fig:soft_inpainting_blending}
\end{figure}

Figure~\ref{fig:soft_inpainting_blending} compares two texturing results: (a) without soft-inpainting, i.e., with the confidence-based blending only, and (b) with both soft-inpainting and confidence-based blending. The results show that both soft-inpainting and confidence-based blending are essential for reducing seams. 


\subsubsection{Performances}

For TEXTure and Text2Tex, we use 8 and 26 viewpoints, respectively, as set in their source codes, whereas 4 viewpoints are used for both Paint3D and RoCoTex. 
Table~\ref{tab:runtime} compares the runtime and memory consumed by RoCoTex and the baselines. RoCoTex runs faster than TEXTure and Text2Tex but is slightly slower than Paint3D. This can be attributed to the difference in inference time between the backbones: original Stable Diffusion models (Paint3D) and SDXL (RoCoTex). On the other hand, RoCoTex requires more memory than the others, due to the inference resolution.

\begin{table}
\setlength{\tabcolsep}{8pt}
    \caption{Runtime and memory consumption.}
    \centering
    \begin{tabular}{ccc}
    \toprule
    Method  & Runtime (sec) & Memory (GB) \\
    \midrule
    TEXTure  & 94 & 12.2  \\
    \cmidrule(lr){1-1} \cmidrule(lr){2-2} \cmidrule(lr){3-3}
    Text2Tex  & 446 & 12.8  \\
    \cmidrule(lr){1-1} \cmidrule(lr){2-2} \cmidrule(lr){3-3}
    Paint3D & 36 & 12.8 \\
    \midrule
    RoCoTex  &  55 & 24.1 \\
    \bottomrule
    \end{tabular}
    \label{tab:runtime}
\end{table}

\subsection{Limitation and Future Work}

Despite the advancements, our approach does not fully resolve occlusion issues due to the limitations of the iterative texture synthesis strategy. 
Although using many views can somewhat mitigate the issue, there can still be angles that the diffusion model fails to generate well, and there is also a trade-off with speed. 
This issue will be addressed in future work. 
Additionally, our research did not address lighting issues, but we found that training a 2D diffusion model can somewhat mitigate these lighting challenges. 
This will also be addressed in future work.
Although the current extrapolation strategy fills in the untextured areas, it may not always produce accurate results.
Future work could investigate the generation process in UV space to synthesize textures for occluded regions.

Furthermore, our method suffered from the baked-in illumination problem, where the generated textures may include unwanted lighting information.
we will investigate methods to disentangle intrinsic material properties from the lighting information during the texture generation process.

\begin{figure*}[ht]
    \centering
    \includegraphics[width=0.9\linewidth]{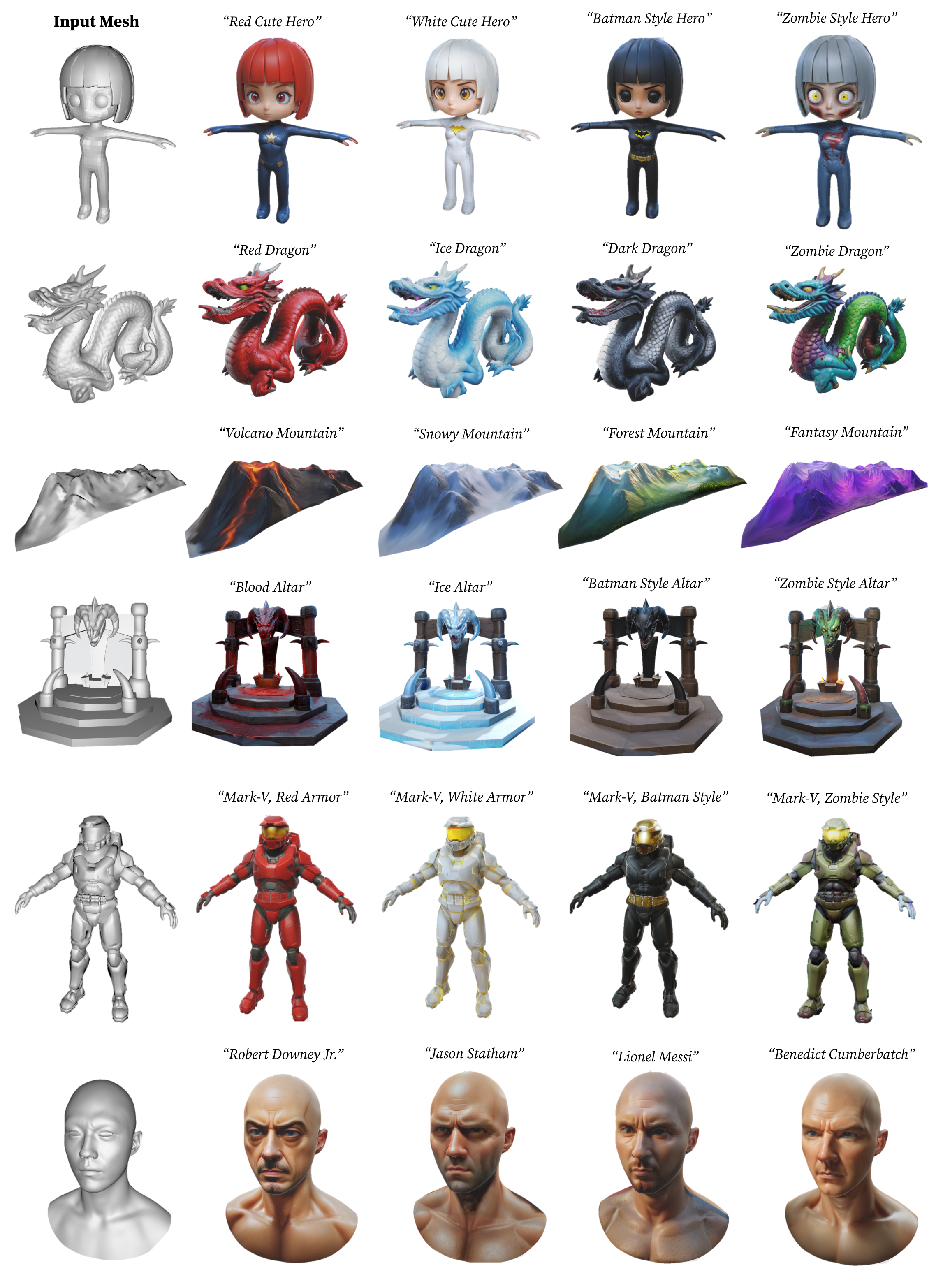}
    \caption{Texturing results with various text prompts.}
    \label{fig:variation}
\end{figure*}


\section{Discussion and Conclusion}
\label{sec:conclusion}

This paper presents RoCoTex, a novel method for generating high-quality consistent textures in a robust manner. It addresses the challenges encountered by existing texturing techniques through symmetrical view synthesis, regional prompting, and  integration of SDXL and multiple ControlNets. Incorporating Differential Diffusion-based soft-inpainting and confidence-based texture blending further enhances the seamlessness and visual integrity of the generated textures.
Experimental results demonstrate the effectiveness and superiority of RoCoTex.

Despite the advancements, RoCoTex does not fully resolve the occlusion issue due to the limitations of the iterative texture synthesis strategy. 
Although using many views can somewhat mitigate the issue, there can still be angles that the diffusion model fails to handle properly, and there is also a trade-off with speed. (In the current implementation, the holes in the texture, which are generated due to occlusion, are filled by interpolation.) 
Additionally, our research does not address lighting issues. 
These challenges will be addressed in future work.

{
    \small
    \bibliographystyle{splncs04}
    \bibliography{main}

\begin{thebibliography}{10}
\providecommand{\url}[1]{\texttt{#1}}
\providecommand{\urlprefix}{URL }
\providecommand{\doi}[1]{https://doi.org/#1}

\bibitem{kid}
Bi{\'n}kowski, M., Sutherland, D.J., Arbel, M., Gretton, A.: Demystifying mmd gans. arXiv preprint arXiv:1801.01401  (2018)

\bibitem{texfusion}
Cao, T., Kreis, K., Fidler, S., Sharp, N., Yin, K.: Texfusion: Synthesizing 3d textures with text-guided image diffusion models. In: Proceedings of the IEEE/CVF International Conference on Computer Vision. pp. 4169--4181 (2023)

\bibitem{text2tex}
Chen, D.Z., Siddiqui, Y., Lee, H.Y., Tulyakov, S., Nie{\ss}ner, M.: Text2tex: Text-driven texture synthesis via diffusion models. In: Proceedings of the IEEE/CVF International Conference on Computer Vision. pp. 18558--18568 (2023)

\bibitem{fantasia3d}
Chen, R., Chen, Y., Jiao, N., Jia, K.: Fantasia3d: Disentangling geometry and appearance for high-quality text-to-3d content creation. In: Proceedings of the IEEE/CVF International Conference on Computer Vision. pp. 22246--22256 (2023)

\bibitem{chen2022tango}
Chen, Y., Chen, R., Lei, J., Zhang, Y., Jia, K.: Tango: Text-driven photorealistic and robust 3d stylization via lighting decomposition. Advances in Neural Information Processing Systems  \textbf{35},  30923--30936 (2022)

\bibitem{chen2022auv}
Chen, Z., Yin, K., Fidler, S.: Auv-net: Learning aligned uv maps for texture transfer and synthesis. In: Proceedings of the IEEE/CVF Conference on Computer Vision and Pattern Recognition. pp. 1465--1474 (2022)

\bibitem{objaverse}
Deitke, M., Schwenk, D., Salvador, J., Weihs, L., Michel, O., VanderBilt, E., Schmidt, L., Ehsani, K., Kembhavi, A., Farhadi, A.: Objaverse: A universe of annotated 3d objects. In: Proceedings of the IEEE/CVF Conference on Computer Vision and Pattern Recognition. pp. 13142--13153 (2023)

\bibitem{ddpm}
Ho, J., Jain, A., Abbeel, P.: Denoising diffusion probabilistic models. Advances in neural information processing systems  \textbf{33},  6840--6851 (2020)

\bibitem{shap-e}
Jun, H., Nichol, A.: Shape: Generating conditional 3d implicit functions. arXiv preprint arXiv:2305.02463  (2023)

\bibitem{soft_inpainting}
Levin, E., Fried, O.: Differential diffusion: Giving each pixel its strength. arXiv preprint arXiv:2306.00950  (2023)

\bibitem{magic3d}
Lin, C.H., Gao, J., Tang, L., Takikawa, T., Zeng, X., Huang, X., Kreis, K., Fidler, S., Liu, M.Y., Lin, T.Y.: Magic3d: High-resolution text-to-3d content creation. In: Proceedings of the IEEE/CVF Conference on Computer Vision and Pattern Recognition. pp. 300--309 (2023)

\bibitem{latent-nerf}
Metzer, G., Richardson, E., Patashnik, O., Giryes, R., Cohen-Or, D.: Latent-nerf for shape-guided generation of 3d shapes and textures. In: Proceedings of the IEEE/CVF Conference on Computer Vision and Pattern Recognition. pp. 12663--12673 (2023)

\bibitem{text2mesh}
Michel, O., Bar-On, R., Liu, R., Benaim, S., Hanocka, R.: Text2mesh: Text-driven neural stylization for meshes. In: Proceedings of the IEEE/CVF Conference on Computer Vision and Pattern Recognition. pp. 13492--13502 (2022)

\bibitem{regional_prompt}
hako mikan: Regional prompter. \url{https://github.com/hako-mikan/sd-webui-regional-prompter} (2023)

\bibitem{clipmesh}
Mohammad~Khalid, N., Xie, T., Belilovsky, E., Popa, T.: Clip-mesh: Generating textured meshes from text using pretrained image-text models. In: SIGGRAPH Asia 2022 conference papers. pp.~1--8 (2022)

\bibitem{texture_field}
Oechsle, M., Mescheder, L., Niemeyer, M., Strauss, T., Geiger, A.: Texture fields: Learning texture representations in function space. In: Proceedings of the IEEE/CVF International Conference on Computer Vision. pp. 4531--4540 (2019)

\bibitem{sdxl}
Podell, D., English, Z., Lacey, K., Blattmann, A., Dockhorn, T., M{\"u}ller, J., Penna, J., Rombach, R.: Sdxl: Improving latent diffusion models for high-resolution image synthesis. arXiv preprint arXiv:2307.01952  (2023)

\bibitem{dreamfusion}
Poole, B., Jain, A., Barron, J.T., Mildenhall, B.: Dreamfusion: Text-to-3d using 2d diffusion. arXiv preprint arXiv:2209.14988  (2022)

\bibitem{dalle2}
Ramesh, A., Dhariwal, P., Nichol, A., Chu, C., Chen, M.: Hierarchical text-conditional image generation with clip latents. arXiv preprint arXiv:2204.06125  \textbf{1}(2), ~3 (2022)

\bibitem{dalle}
Ramesh, A., Pavlov, M., Goh, G., Gray, S., Voss, C., Radford, A., Chen, M., Sutskever, I.: Zero-shot text-to-image generation. In: International conference on machine learning. pp. 8821--8831. Pmlr (2021)

\bibitem{TEXTure}
Richardson, E., Metzer, G., Alaluf, Y., Giryes, R., Cohen-Or, D.: Texture: Text-guided texturing of 3d shapes. In: ACM SIGGRAPH 2023 Conference Proceedings. pp. 1--11 (2023)

\bibitem{ldm}
Rombach, R., Blattmann, A., Lorenz, D., Esser, P., Ommer, B.: High-resolution image synthesis with latent diffusion models. In: Proceedings of the IEEE/CVF conference on computer vision and pattern recognition. pp. 10684--10695 (2022)

\bibitem{laion-5b}
Schuhmann, C., Beaumont, R., Vencu, R., Gordon, C., Wightman, R., Cherti, M., Coombes, T., Katta, A., Mullis, C., Wortsman, M., et~al.: Laion-5b: An open large-scale dataset for training next generation image-text models. Advances in Neural Information Processing Systems  \textbf{35},  25278--25294 (2022)

\bibitem{laion-400m}
Schuhmann, C., Vencu, R., Beaumont, R., Kaczmarczyk, R., Mullis, C., Katta, A., Coombes, T., Jitsev, J., Komatsuzaki, A.: Laion-400m: Open dataset of clip-filtered 400 million image-text pairs. arXiv preprint arXiv:2111.02114  (2021)

\bibitem{mvdream}
Shi, Y., Wang, P., Ye, J., Long, M., Li, K., Yang, X.: Mvdream: Multi-view diffusion for 3d generation. arXiv preprint arXiv:2308.16512  (2023)

\bibitem{texturify}
Siddiqui, Y., Thies, J., Ma, F., Shan, Q., Nie{\ss}ner, M., Dai, A.: Texturify: Generating textures on 3d shape surfaces. In: European Conference on Computer Vision. pp. 72--88. Springer (2022)

\bibitem{ddim}
Song, J., Meng, C., Ermon, S.: Denoising diffusion implicit models. arXiv preprint arXiv:2010.02502  (2020)

\bibitem{texro}
Wu, J., Liu, X., Wu, C., Gao, X., Liu, J., Liu, X., Zhao, C., Feng, H., Ding, E., Wang, J.: Texro: Generating delicate textures of 3d models by recursive optimization. arXiv preprint arXiv:2403.15009  (2024)

\bibitem{yeh2024texturedreamer}
Yeh, Y.Y., Huang, J.B., Kim, C., Xiao, L., Nguyen-Phuoc, T., Khan, N., Zhang, C., Chandraker, M., Marshall, C.S., Dong, Z., et~al.: Texturedreamer: Image-guided texture synthesis through geometry-aware diffusion. arXiv preprint arXiv:2401.09416  (2024)

\bibitem{point-UV}
Yu, X., Dai, P., Li, W., Ma, L., Liu, Z., Qi, X.: Texture generation on 3d meshes with point-uv diffusion. In: Proceedings of the IEEE/CVF International Conference on Computer Vision. pp. 4206--4216 (2023)

\bibitem{2023paint3}
Zeng, X.: Paint3d: Paint anything 3d with lighting-less texture diffusion models. arXiv preprint arXiv:2312.13913  (2023)

\bibitem{controlnet}
Zhang, L., Rao, A., Agrawala, M.: Adding conditional control to text-to-image diffusion models. In: Proceedings of the IEEE/CVF International Conference on Computer Vision. pp. 3836--3847 (2023)

\bibitem{zhang2024mapa}
Zhang, S., Peng, S., Xu, T., Yang, Y., Chen, T., Xue, N., Shen, Y., Bao, H., Hu, R., Zhou, X.: Mapa: Text-driven photorealistic material painting for 3d shapes. arXiv preprint arXiv:2404.17569  (2024)

\end{thebibliography}
}

\end{document}